\documentclass[10pt,journal,compsoc]{IEEEtran}
\ifCLASSOPTIONcompsoc
  \usepackage[nocompress]{cite}
\else
  \usepackage{cite}
\fi

\usepackage{graphicx}
\usepackage{subfigure}
\usepackage{amssymb}
\usepackage[ruled]{algorithm2e}
\usepackage{color}
\usepackage{ragged2e}
\usepackage{cite}

\usepackage{hyperref}
\usepackage{multirow}
\usepackage{colortbl}
\usepackage{amsmath}
\usepackage{pifont}
\usepackage{balance}
\usepackage{booktabs}
\usepackage{xcolor}

\definecolor{darkgreen}{rgb}{0.0, 0.2, 0.13}

\usepackage{diagbox}

\ifCLASSINFOpdf
\else
\fi
\usepackage{algorithmic}
\begin{document}
%
\title{
JM3D \& JM3D-LLM: Elevating 3D Understanding with Joint Multi-modal Cues
}
%
%
%
%

\author{
        Jiayi Ji,
        Haowei Wang,
        Changli Wu,
        Yiwei Ma, 
        Xiaoshuai Sun,
        Rongrong Ji

\IEEEcompsocitemizethanks{

\IEEEcompsocthanksitem J. Ji, H. Wang, C. Wu, Y. Ma, X. Sun and R. Ji are with Key Laboratory of Multimedia Trusted Perception and Efficient Computing, Ministry of Education of China, Xiamen University, 361005, P.R. China. \protect
\IEEEcompsocthanksitem J. Ji and H. Wang contributed equally to this work.
}

}

\IEEEtitleabstractindextext{%
\begin{abstract}
\justifying
The rising importance of 3D understanding, pivotal in computer vision, autonomous driving, and robotics, is evident. However, a prevailing trend, which straightforwardly resorted to transferring 2D alignment strategies to the 3D domain, encounters three distinct challenges: (1) Information Degradation: This arises from the alignment of 3D data with mere single-view 2D images and generic texts, neglecting the need for multi-view images and detailed subcategory texts. (2) Insufficient Synergy: These strategies align 3D representations to image and text features individually, hampering the overall optimization for 3D models. (3) Underutilization: The fine-grained information inherent in the learned representations is often not fully exploited, indicating a potential loss in detail. To address these issues, we introduce JM3D, a comprehensive approach integrating point cloud, text, and image. Key contributions include the Structured Multimodal Organizer (SMO), enriching vision-language representation with multiple views and hierarchical text, and the Joint Multi-modal Alignment (JMA), combining language understanding with visual representation. Our advanced model, JM3D-LLM, marries 3D representation with large language models via efficient fine-tuning. Evaluations on ModelNet40 and ScanObjectNN establish JM3D's superiority. The superior performance of JM3D-LLM further underscores the effectiveness of our representation transfer approach. Our code and models are available at~\url{https://github.com/Mr-Neko/JM3D}.

\end{abstract}

\begin{IEEEkeywords}
3D Understanding, Structured Multimodal Organizer, Joint Multi-modal Alignment, Large Language Model.
\end{IEEEkeywords}}

\maketitle

\IEEEdisplaynontitleabstractindextext

%
\IEEEpeerreviewmaketitle

\IEEEraisesectionheading{\section{Introduction}\label{sec:Introduction}}

%
%
%
%
\IEEEPARstart
The study of 3D model understanding~\cite{achlioptas2018learning, liu2019densepoint, liu2020closer, ran2022surface, xie2020grnet, xu2021paconv, wang2022p2p} has gained prominence, especially in areas like augmented/virtual reality~\cite{liu2021group, vu2022softgroup, armeni20163d} and autonomous driving~\cite{li2022deepfusion, yin2021center}. However, limited data availability and suboptimal category representation pose challenges for 3D understanding, especially when compared to the abundance of image-text pair data.

Addressing the challenge of limited 3D data, recent studies~\cite{hegde2023clip, xue2022ulip, zhang2023clip} have leveraged the rich resources of other modalities. They utilize large-scale vision-language models like CLIP~\cite{radford2021learning} to enhance 3D representation. The core idea is to integrate 3D features into the combined vision-language space, leveraging the robust zero-shot capabilities of foundational models. Commonly, an image rendered from a specific view of a 3D model, paired with a basic category label, is inputted into CLIP. The 3D features are subsequently aligned to the vision-language domain through a contrastive methodology. As proven by ULIP~\cite{xue2022ulip} and CG3D~\cite{hegde2023clip}, this approach, enriched by external information, markedly bolsters 3D understanding and showcases commendable transferability.

\begin{figure}[]
\centering
\includegraphics[width=1.0\columnwidth]{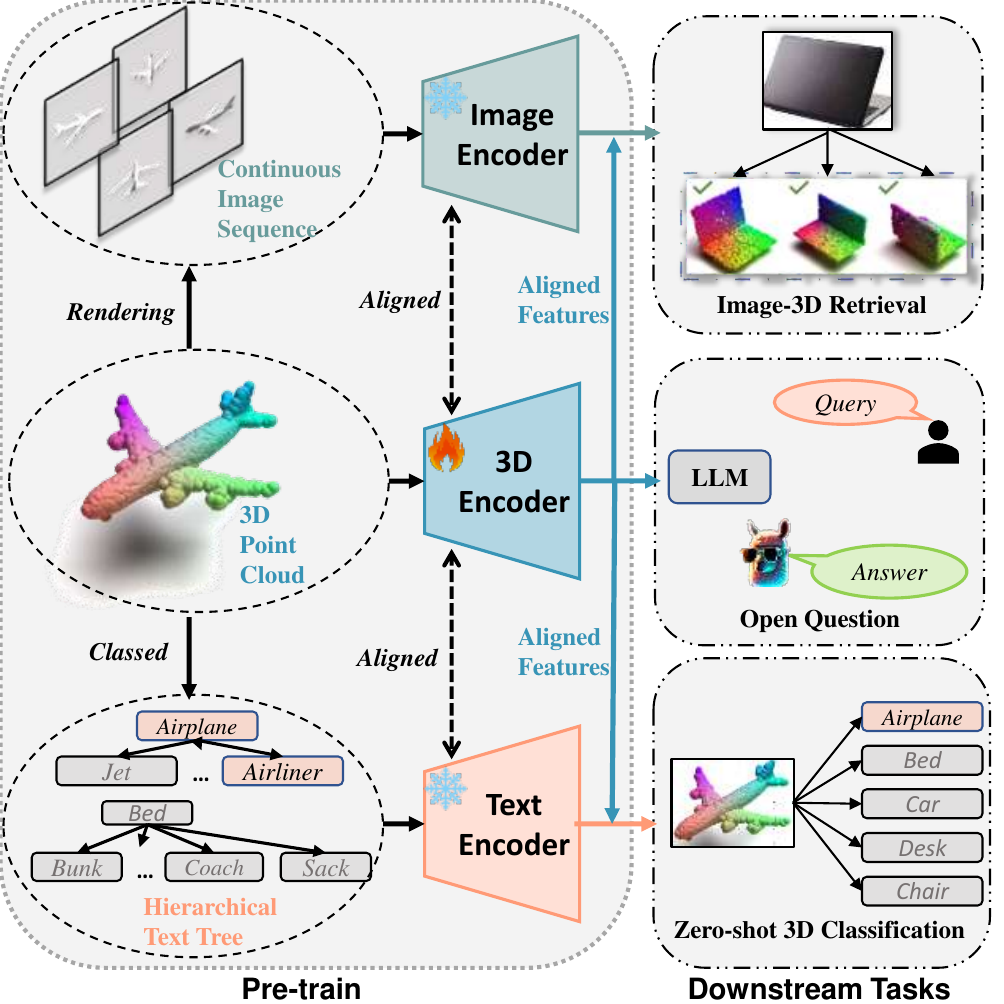}
\caption{The visualization of JM3D. JM3D coherently aligns the 3D modality with previously aligned vision and language modalities, forming a consolidated tri-modal representation. Subsequently, the derived representation finds application in tasks such as image-3D retrieval, zero-shot 3D classification, and interfaces with LLM to discern more granular information. 
}
\label{fig1}
\end{figure}

While prevalent methods~\cite{hegde2023clip, xue2022ulip, zhang2023clip} lean on 2D alignment techniques for 3D representation learning, they often overlook the inherent complexities of 3D models. This oversight results in three principal challenges: 
(1) \textbf{Information Degradation}: By aligning 3D models with single-view images and broad text descriptions, essential spatial and depth details are lost. For example, in single-view images such as Fig.~\ref{fig1}, the airplane's front render misses out on wing specifics, much like its rear view. Textually speaking, a generalized label like ``airplane'' doesn't provide distinctions between specific aircraft categories, like airliners versus jets or bombers. 
(2) \textbf{Insufficient Synergy}: These methods align 3D representations with image features and text features separately, neglecting the joint modeling of vision and language modalities. This issue complicates the optimization process for 3D representations, making it difficult to determine whether to move closer to image features or text features, and to what extent, leading to incomplete information utilization. 
(3) \textbf{Underutilization}: Moreover, the learned representations frequently fall short in harnessing and further cultivating the available granular details. Successful transfer from 2D representations should theoretically endow the 3D model not just with categorical insights but also with finer attributes and subtler characteristics.

To address the above challenges, we introduce JM3D, a novel multi-modal approach for comprehensive 3D representation learning, as showcased in Fig.~\ref{fig1}. JM3D pivots on two primary modules: the Structured Multi-modal Organizer (SMO) and the Joint Multi-modal Alignment (JMA). We first propose SMO to address information degradation, which enhances each modality independently. For visual enhancement, we assert that a 3D model should correlate with a continuous array of images spanning diverse angles. Consequently, a Continuous Image Sequence (CIS) is introduced to concurrently model multiple viewpoints. To capture intricate details, we encode attributes such as angle, color, and depth into these images. 
From a linguistic perspective, our Hierarchical Text Tree (HTT) augments textual representation. By incorporating specific sub-categories, like ``jet'', ``airliner'', or ``bomber'', the system gains granularity in understanding. Meanwhile, coarser classifications like ``airplane'' consolidate semantically related sub-categories, reinforcing the robustness of our model.
We then design JMA to tackle the insufficient synergy. JMA synergistically aligns both visual and language modalities, engaging them concurrently during optimization. This synergy ensures the 3D model effectively harnesses insights from both domains. Furthermore, we provide a theoretical framework underscoring our approach's effectiveness, laying the groundwork for forthcoming explorations in the domain. 
%
%
Building upon the feature alignment accomplished by our JM3D framework, the 3D representation seamlessly synchronizes with textual descriptions, thereby setting the stage for its potential integration into the Large Language Model (LLM). Motivated by the well-regarded technique of Instruction Tuning, we employ Parameter-Efficient fine-tuning to embed 3D semantics into the LLM. This fusion culminates in JM3D-LLM, enabling the LLM to parse more granular cues, thereby addressing the underutilization issue. To anchor this integration, we've curated a conversational dataset from Cap3D tailored for training.

Our comprehensive experiments on both ModelNet40 and ScanObjectNN underscore the potency of our introduced approach, JM3D, which sets a new benchmark in zero-shot 3D classification. In particular, when compared to ULIP, JM3D demonstrates a substantial performance advantage. It surpasses PointMLP by an impressive margin of approximately 4.3\%, and achieves a remarkable increase in accuracy of up to 6.5\% when evaluated with PointNet++ for top-1 accuracy in zero-shot 3D classification on the ModelNet40 dataset. The results achieved with JM3D-LLM further reinforce the strength and adaptability of our representation. The nuanced and detailed descriptions generated by JM3D highlight its proficiency in discerning and capturing intricate features.

In summary, this paper presents three key contributions:
\begin{itemize}
    \item To combat the challenge of information degradation, we introduce the Structured Multimodal Organizer (SMO). This framework generates a continuous sequence of multi-view rendered images and establishes a hierarchical text tree. Through the augmentation of both visual and textual modalities, the SMO effectively offsets the diminishing 3D visual attributes, thereby facilitating a richer and more encompassing representation.
    
    \item To address the challenge of inadequate synergy, we introduce the Joint Multi-modal Alignment (JMA). This method seamlessly integrates textual and visual modalities, leading to a unified representation. By doing so, the JMA minimizes potential suboptimal outcomes and enhances the coherent interpretation of image-text pairings.

    \item We introduce JM3D-LLM, which seamlessly embeds 3D representations into the Large Language Model (LLM) using an efficient training method. Evaluation on ModelNet40 and ScanObjectNN highlight JM3D's excellence in zero-shot 3D classification. The exciting results achieved by JM3D-LLM further emphasize JM3D's robust transfer learning capabilities and its ability to discern important features.
\end{itemize}

\section{Related work}
\subsection{Representation Learning in 3D Space}
3D representation learning seeks to derive semantic features of 3D models. Among various representation techniques for these models, point clouds have emerged as a favored input format in deep learning due to their sparsity and discreteness~\cite{aubry2011wave, bronstein2010scale, sun2009concise, wu20153d, maturana2015voxnet, zhao-etal-2023-generating-visual, wang2022p2p}. Initial approaches~\cite{maturana2015voxnet, shi2020pv} relied on extracting point cloud data from voxels and applying convolutions for global feature capture. However, subsequent strategies, such as PointNet~\cite{qi2017pointnet}, PointNext~\cite{qian2022pointnext}, and PointMLP~\cite{marethinking}, devised architectures tailored for point clouds. Particularly, PointNet pioneered the extraction of permutation-invariant features from point clouds, laying groundwork for future methodological designs. The recent PointMLP employs dual MLP blocks and a geometric transformer, achieving notable outcomes without necessitating intricate local geometric feature extractors.

The advent of transformers~\cite{vaswani2017attention} has spurred self-supervised learning methods~\cite{guo2021pct, liu2022masked, xiao2023unsupervised} to generate augmented point clouds. These methods leverage encoder-decoder architectures to reconstruct point clouds, a strategy found effective in models like PointBert~\cite{yu2022point}, Point-MAE~\cite{pang2022masked}, and Point-M2AE~\cite{zhangpoint}.

Nevertheless, independent of architectural innovations, the paramount challenge in 3D representation learning remains the small-scale datasets. Many methods grapple with rudimentary category annotations and scarce data, leading to compromised robustness in real-world scenarios.

\subsection{Representation Learning in Multi-modal Space}
Representation learning in multi-modal space mainly aims at aligning semantics features from different modalities. Contemporary techniques largely fall into two categories. The first merges features from various modalities using intricate architectures~\cite{li2019visualbert, li2020oscar, chen2020uniter, fei-etal-2023-scene} and fosters alignment through tasks like retrieval and comprehension. Such methods~\cite{li2019visualbert, li2020oscar, lu2019vilbert, tan2019lxmert, wang2023towards} predominantly delve into the synergy between image regions and their corresponding textual descriptions.

On the other hand, methods exemplified by CLIP~\cite{radford2021learning} lean on contrastive learning across visual and linguistic domains, achieving direct feature alignment with an extensive set of positive and negative examples. Successive techniques have refined CLIP's alignment, considering both the scope of data and its granularity. Specifically, the GLIP series~\cite{li2022grounded, zhang2022glipv2} seeks nuanced alignment through detection-oriented tasks, while Flamingo~\cite{alayrac2022flamingo} employs a more expansive dataset. Concurrently, works~\cite{luo2022clip4clip, li2022align, xu2021videoclip, ju2022prompting, FeiMatchStruICML22} validate the CLIP model's adaptability, underscoring its efficacy even when applied to diverse modalities, including video and text.
 
\subsection{Enhancing 3D Representation through Multi-modality}
The merits of multi-modal representation learning, especially its enhanced performance from assimilating diverse modalities, are becoming increasingly evident. Tapping into the rich knowledge reservoir of image-text pre-trained models to augment 3D model representations offers promising avenues~\cite{chen2021multimodal, yanlet, ma2023xmesh, wang2023beyond}. Pioneering this approach, PointCLIP series~\cite{zhang2022pointclip, zhu2023pointclip} leveraged a visual-language pre-trained model for point cloud models, transforming point clouds into a sequence of depth images which were subsequently input into CLIP for zero-shot classification. However, its potential to truly enhance the IEEEtranTPAMIexpressiveness of point clouds remained untapped. 
ULIP~\cite{xue2022ulip} and CG3D~\cite{hegde2023clip}, in contrast, endeavor to adapt the CLIP paradigm directly, striving for a cohesive representation space encompassing point clouds, text, and images. They deploy a contrastive learning approach, delineating relations between the point cloud modality with both the visual and language modalities independently. Yet, such methods might be shortsighted, as they often equate the richness of a solitary image or text fragment to a comprehensive 3D model.  Furthermore, this segmentation fails to capture the joint distribution intrinsic to image-text combinations. In parallel developments, ULIP2~\cite{xue2023ulip} approached the challenge from a data-centric perspective, generating extensive descriptions for images from each viewpoint, significantly augmenting the scalability and comprehensiveness of 3D representations. In contrast, our work focuses on enhancing capabilities through innovative training strategies. The contributions of both methodologies are thus orthogonal and complementary.
%

\begin{figure*}[]
\centering
\includegraphics[width=1.0\textwidth]{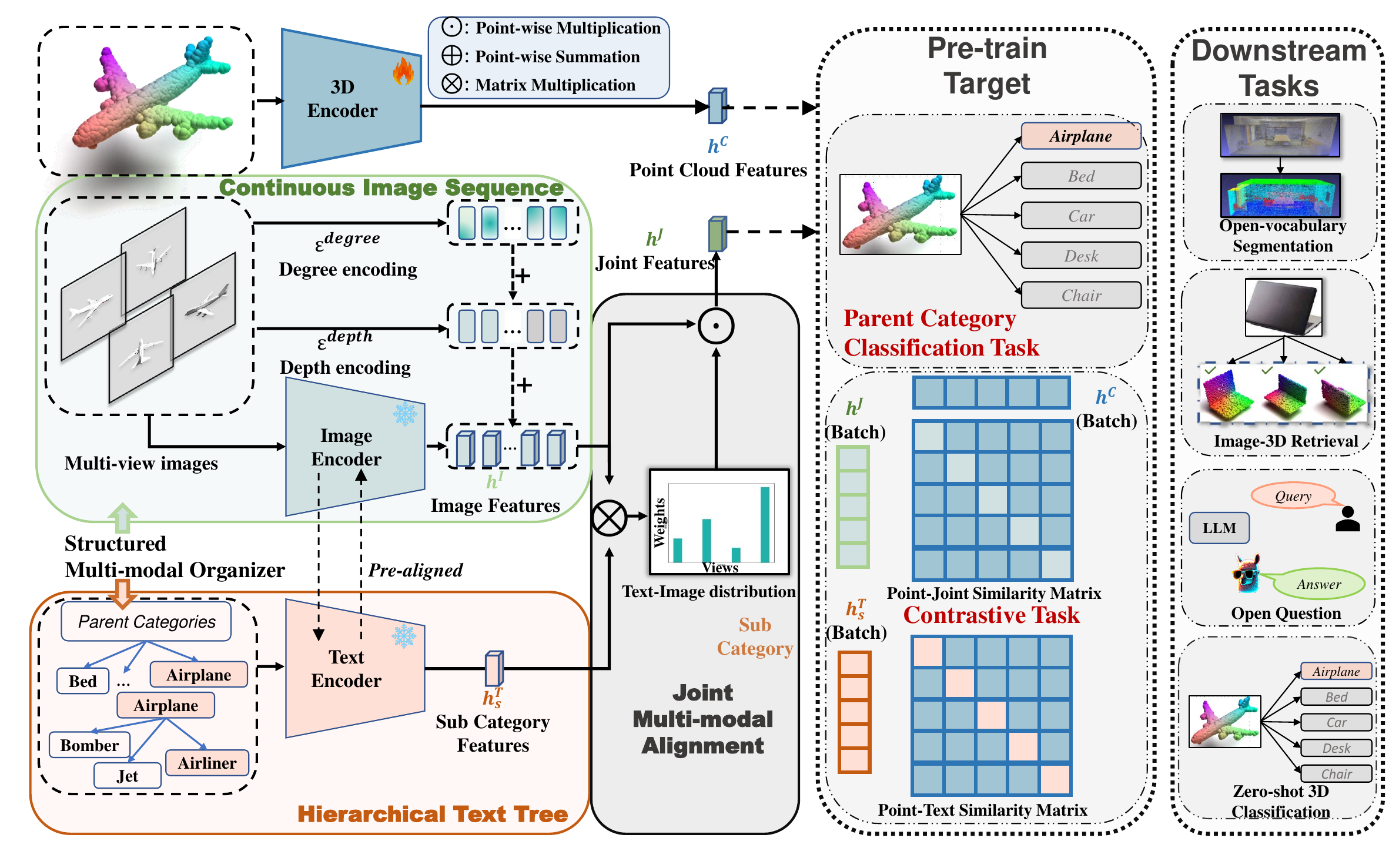}

\caption{The framework of JM3D. Continuous Image Sequence {(CIS)} and Hierarchical Text Tree {(HTT)} organized continuous multi-view images and hierarchical texts respectively, which are fed into a pre-training model (frozen) to extract features on the left. Then, Joint Multi-modal Alignment {(JMA)} incorporates the features from two modalities to generate the joint modeling features. On the last, contrastive learning is applied to align 3D features (training) with joint features and subcategory texts, while 3D features are aggregated with the assistance of the parent category.}
\label{fig2}
\end{figure*}

\subsection{Large Language Model}
In contemporary natural language processing (NLP) research, the emphasis has increasingly shifted towards Large Language Models (LLM), characterized by their expansive parameter counts and intensive pre-training on vast datasets. The efficacy of LLMs is evident in their superior performance across varied linguistic tasks such as translation, reasoning, and conversational engagement. The GPT series~\cite{radford2018improving, radford2019language, radford2021learning} stands as a testament to this trend, consistently enhancing both model size and training data breadth. In the quest for optimal data utilization, the concept of prompt learning~\cite{liu2023pre, houlsby2019parameter, petroni2019language, schick2021exploiting} emerged, facilitating a streamlined approach to diverse task formats. This has paved the way for models like Instruct GPT~\cite{ouyang2022training}, which boasts a staggering 175 billion parameters. Building on this foundation, techniques such as Supervised Fine-tuning (SFT) and Reinforcement Learning from Human Feedback (RLHF) have been introduced, refining LLM outputs to more closely resemble human-like interactions, as seen in ChatGPT and LLaMA~\cite{touvron2023llama}. The landscape further evolved with multi-modal approaches like MiniGPT-4~\cite{zhu2023minigpt}, LLaVA~\cite{liu2023visual}, and LLaMA-adapter~\cite{zhang2023llama}, leveraging LLM's inferential prowess to decipher visual content. This progression signals an intriguing trajectory, harnessing LLMs to amalgamate 3D data for enriched comprehension. Concurrently, both POINTLLM~\cite{xu2023pointllm} and Point-LLM~\cite{guo2023point} have also delved into this direction. Compared to their efforts, our work specifically aims to investigate the benefits of finer-grained 3D representations in aiding the understanding capabilities of LLMs.

\section{Joint Multi-modal 3D Representation Learning}

We first provide an overview of the contrastive learning framework in Sec.~\ref{sec:ulip}, highlighting its key components and limitations. Next, we delve into the details of our proposed SMO in Sec.~\ref{sec:SMO}. Then, we introduce the JMA in Sec.~\ref{sec:JMA}, focusing on its ability to optimize the aligning process of the model. Finally, we outline the objectives of the construction of unified triplet modality modeling in Sec.~\ref{sec:loss}. See Fig.~\ref{fig2} for more details about the framework.

\subsection{Preliminary \label{sec:ulip}}

ULIP~\cite{xue2022ulip} explores the transfer ability of 2D contrastive learning to 3D by constructing a dataset combining point clouds, rendered images, and language descriptions from ShapeNet55 \cite{chang2015shapenet}. For each CAD model, a triplet sample $S_i: (I_i, T_i, C_i)$ is created, consisting of a rendered image $I_i \in \mathbb{R}^{H \times W \times 3}$, a text description $T_i$, and a sampled point cloud $C_i \in \mathbb{R}^{N_c}$, where $H$, $W$, and $N_c$ represent the height, width of the image, and the number of sampled points, respectively.

Specifically, The image $I_i$ undergoes rendering at a random angle, whereas the description $T_i$ is generated by combining a fixed prompt with a broad category, \emph{i.e.}, ``a 3D representation of [CLASS]''. To accommodate various point cloud backbones, the point cloud $C_i$ is evenly sampled. The objective is to bring together the representations of the triplet modalities in a cohesive semantic space, which can be stated as follows:
\begin{equation}
    P(C, I, T)=P(C|I, T) \cdot P(I, T).
    \label{eq:definition}
\end{equation}
In prior research~\cite{xue2022ulip, hegde2023clip}, Eq.~\ref{eq:definition} has been simplified based on an approximate assumption, which posits the \textit{\textbf{conditional independence of the vision information $I$ and language modality $T$}}. Consequently, the joint modality conditional probability $P(C|I, T)$ is overlooked and the simplification of aligning each modality individually in ULIP can be outlined below:
\begin{equation}
    P(C, I, T)=P(C|I) \cdot P(C|T) \cdot P(I, T).
    \label{eq:ULIP}
\end{equation}
Drawing inspiration from Eq.~\ref{eq:ULIP}, although ULIP adopts contrastive learning to effectively learn the joint distribution of trimodal, it focuses on aligning the 3D features with language and vision features separately. Specifically, the joint probability distribution $P(I, T)$ can be obtained by employing a pre-trained vision-language model, such as CLIP\cite{radford2021learning}. This pre-trained model serves as a valuable resource for pre-aligned features, while various backbones~\cite{marethinking, qi2017pointnet++, yu2022point} extract the 3D features. These operations can be formulated as the following:
\begin{equation}
    h_i^{C},\ h_i^I,\ h_i^T=f_{C}({C}_i),\ f_I(I_i),\ f_T(T_i),
    \label{eq:extract}
\end{equation}
where $f_{C}$ is a 3D backbone. $f_I$ and $f_T$ are the vision encoder and language encoder from the pre-trained vision-language model,  adopting vanilla transformer~\cite{vaswani2017attention} structures. The vectors $h_i^C\in \mathbb{R}^{D}$, $h_i^I\in \mathbb{R}^{D}$, and $h_i^T\in \mathbb{R}^{D}$ denote 3D, language, and vision features, respectively, and $D$ represents the dimensions of the final representation vectors.

A contrastive way~\cite{radford2021learning} followed is chosen to construct the conditional distribution between any two modalities in Eq.~\ref{eq:ULIP}, aligning $h_i^C$, $h_i^I$, and $h_i^T$ by:
\begin{equation}
 \mathcal{L}_{(h^{M_1}, h^{M_2})} =
   \sum_i-\frac{1}{2}log\frac{\exp\left({\mathbf{h}^{M_1}_i \mathbf{h}^{M_2}_i}\right)}{\sum_k \exp\left({\mathbf{h}^{M_1}_i \mathbf{h}^{M_2}_k}\right)},
    \label{eq:ULIP contrastive}
\end{equation}
where $M = (M_1, M_2) \in \{(T, I), (C, T), (C, I)\}$ represents the combination of pairwise modalities. 

\subsection{Structured Multi-modal Organizer \label{sec:SMO}}
Structured Multi-modal Organizer (SMO) is a data-refined module to fill the gap of information between the 3D model $C_i$ with the image $I_i$ and text $T_i$. 
For example, from a visual perspective, consider a 3D model of a car. A single frontal rendered image of the car may not capture crucial information about the rear end of the vehicle. Similarly for the language aspect, using the term ``bottle'' does not accurately represent specific models like ``jug'', ``beer'', or ``flask''. To alleviate this information loss, we adopt a multi-view approach to organize the data in the triplet sample $S_i$, constructing a refreshed form of triplet and redefining it as follows:
\begin{equation}
    S_i:\left(\left[I_{i1}, \cdots, I_{iv}\right], \left[T_{i}^p, T_{i}^s\right], C_i\right).
    \label{eq:sample}
\end{equation}
To ensure a more fair and comprehensive alignment between the trimodal, SMO incorporates $v$ images from Continuous Image Sequence (CIS) and structured texts from the Hierarchical Text Tree (HTT). By leveraging these additional visual and textual cues, we capture more accurate and detailed associations between the 3D models, visual representations, and the accompanying textual descriptions.

\subsubsection{Continuous Image Sequence}
When considering the visual modality, the single synthetic image captured from random angles only provides a partial representation of a 3D feature. It is imperative to introduce a set of multi-view rendered images to enhance the semantic information.

More clearly, we synthesize RGB and depth images at regular intervals of 12 degrees, resulting in a candidate image set denoted as $C_I \in \mathbb{R}^{30 \times 3}$. However, we have observed that significant angular deviations between consecutive sampled images $\left[I_{1}, \cdots, I_{v}\right]$ may lead to training instability due to the discrete nature of the sampling process. To overcome this challenge, we employ a novel approach where, during each training process, we selectively sample $v$ images within a specific angular range. This meticulous strategy ensures a more stable and effective training process. The process can be formulated as:
\begin{equation}
\left[I_{1}, \cdots, I_{v}\right] = WS(C_I), |\angle I_i-\angle I_j|<\omega, \forall i, j \in [1, v],
\end{equation}
where $WS(\cdot)$ means sampling within a specific angular range, $\angle I_i$ denotes the render degree of the $i$-th image, and $\omega$ is a hyperparameter set to $60^\circ$ based on our experiments.

The image encoder embeds the multi-view images into the feature vectors. Additionally, considering the significance of angle and depth as positional information within images, we employ angle and depth encodings, inspired by the approach described in \cite{vaswani2017attention}, to capture the spatial information inherent. As depicted in Fig.~\ref{fig2}, these encodings are combined with the visual features through an addition operation, yielding the following formulation:
\begin{equation}
\small
h_{iv}^{I}=\text{LayerNorm}\left(f_I(I_{iv})+\epsilon^{degree}[\angle I_i]+\epsilon^{depth}[\angle I_i]\right),
\end{equation}
where $\text{LayerNorm}(\cdot)$~\cite{1607.06450} controls the range of vision vectors $h_{iv}^{I}$, consequently expediting convergence.

\subsubsection{Hierarchical Text Tree}
When considering the language modality, it is worth noting that existing methods such as ULIP solely rely on simple parent categories (55 categories). While this approach facilitates the training target, it also limits its robustness. However, the ShapeNet55 dataset~\cite{chang2015shapenet} provides subcategories (205 categories) that offer more detailed distinctions. By introducing subcategories, the model can focus on capturing finer-grained representations and accurately differentiate visually similar subcategories.

Nevertheless, when working with limited dataset sizes, independently incorporating subcategories may overlook the family relationships. These relationships are crucial as they indicate that features of subcategories belonging to the same parent category should be more similar. To address this issue, we propose a novel approach called Hierarchical Text Tree (HTT) that constructs a hierarchical category tree for each triplet. This tree comprises coarse semantic parent categories and more specific subcategories (e.g., ``bed'' $\rightarrow$ ``bunk''). In cases where subcategory annotations are incomplete, the parent category is used as a replacement. By employing HTT, we assign structured category information to each model as $[T^p, T^s]$, where $T^p$ represents the parent category with primary semantics, and $T^s$ represents the subcategory with fine-grained details.

To leverage the hierarchical-grained categories, we design specific tasks for both parent and subcategories. The subcategories $T^s$ undergo the process described in Eq.~\ref{eq:extract}, utilizing the language encoder to generate text features that are learned in a contrastive manner. Additionally, the parent categories guide the aggregation of point cloud features, as expressed by the following equation:
\begin{equation}
\mathop{\arg\min}\limits_{\theta} g(\theta)={||\theta(h_{i}^{C})-f(T_{i}^{P})||}^2,
\label{eq:parent classed}
\end{equation}
where $\theta$ is an MLP~\cite{tolstikhin2021mlp}, and $f(\cdot)$ transfers each parent category annotation of $i$-th sample to a numbered label. This approach strikes a balance between specific subcategories and their broader parent categories, enabling the model to leverage both detailed distinctions and overarching similarities within the hierarchical category structure. Consequently, the model becomes more robust, capable of capturing intricate nuances and generalizing well across different instances within the given dataset.

\begin{figure*}[]
\centering
\includegraphics[width=0.96\textwidth]{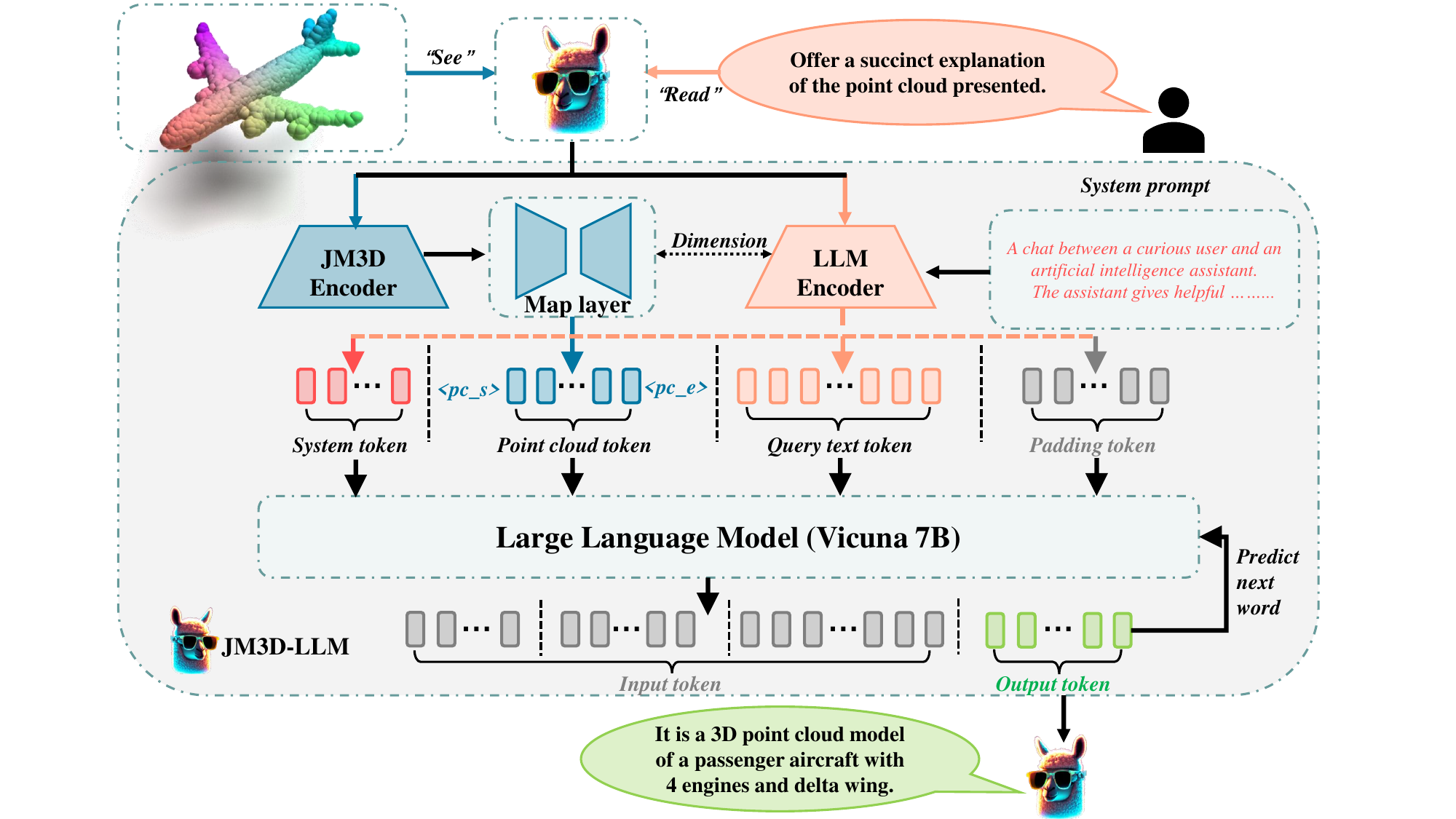}

\caption{The framework of JM3D-LLM. We take the LLM as the cornerstone to support the further semantic understanding task like the fine-grained 3D model captioning.}
\label{fig4}
\end{figure*}

\subsection{Joint Multi-modal Alignment \label{sec:JMA}}
In previous methods, the assumption of conditional independence between the vision and language modalities was described as Eq.~\ref{eq:ULIP}. However, this approximation fails to fully capture the synergistic relationship between the two modalities as Eq.~\ref{eq:definition}, leading to \textbf{\emph{insufficient synergy}} issue, which is the main reason of suboptimal performance. To address this limitation and unlock the potential of the model, we propose a more comprehensive approach called Joint Multi-modal Alignment (JMA) that directly models the relationships between the vision and language modalities, denoted as $P(C | I, T)$. This reformulated relationship can be expressed as:
\begin{equation}
\small
\begin{aligned}
    P(C|I, T)&=\frac{P(C, I|T)}{P(I|T)}=\frac{P(C, I|T)P(T)}{P(I|T)P(T)}=\frac{P(C, I|T)P(T)}{P(I, T)} \\
    &\propto \sum_{i}P(C, I|T_{i})P(T_{i})=\sum_{i}\sum_{j}P(C, I_{i, j}|T_{i})P(T_{i}),
    \label{eq:joint}
\end{aligned}
\end{equation}
that means that for any sample $S_i$ in Eq.~\ref{eq:sample}, both the language and multi-view vision information should be incorporated to take part in the process of contrastive process. Specifically, JMA generates multiple image features $h_{iv}^{I} \in \mathbb{R}^{V \times D}$ and fine-grained text feature $h_{is}^{T}\in \mathbb{R}^{D}$. A similarity matrix between the two features are caluated as the weight to facilitate the reconstruction of image features into joint modality features. This process can be expressed as follows:
\begin{equation}
    h_i^J=\sum_v^V\text{Softmax}(h_{iv}^{I} \times h_{is}^{T}) \otimes h_{iv}^{I},
\end{equation}
where $\times$ denotes matrix multiplication, and $\otimes$ is element-wise multiplication.

The incorporation of JMA allows the model to capture the underlying connections and semantic associations between vision and language modalities, resulting in the representation of the joint vision-language modality. This representation is designed to align more closely with the distribution of the joint semantic space, facilitating the convergence of 3D features into the same space during the process of comparative learning.

\begin{table}[]
\centering
    \caption{Instruction conversation constructing. The Point means the position where to inject the features. }
    \begin{tabular}{l}
        \toprule
        \{Large Language Model Prompt\} \\
        \midrule
        USER:         \textcolor{blue}{$\langle$ point $\rangle$ \textbackslash n} \color{red}\{Random Instruction\} \\
        ASSISTANT: \color{green}\{Caption\}$\langle$ \textbackslash s $\rangle$ \\
        \midrule
        USER:           \textcolor{red}{\{Random Instruction\}} \textcolor{blue}{$\langle$ point $\rangle$ \textbackslash n} \\
        ASSISTANT: \color{green}\{Caption\}$\langle$ \textbackslash s $\rangle$ \\
        \bottomrule
    \end{tabular}
    \label{tab:instruct}
\end{table}

\subsection{Training Objective \label{sec:loss}}

During the training process, we establish two tasks for unified 3D representation learning. The first task involves utilizing a contrastive approach between the point cloud features $h^{C}$ and the other features $h^{J}$ and $h_s^T$, as exemplified in Eq.~\ref{eq:ULIP contrastive}. This task is formulated as follows:
\begin{equation}
    \mathcal{L}_{contrastive}=\lambda_1\mathcal{L}_{(h^{C}, h^J)} + \lambda_2\mathcal{L}_{(h^{C}, h_s^T)} + \lambda_3\mathcal{L}_{(h_s^T, h^J)},
\end{equation}
where the $\lambda_1$, $\lambda_2$, and $\lambda_3$ are the hyper parameters. Through contrastive learning, the features from the three modalities are brought together into a unified semantic space. In this process, the guidance provided by the $h_s^T$ feature plays a crucial role in capturing fine-grained details.

By the way, another task is a classification task with parent category label as Eq.~\ref{eq:parent classed}, which is formulated as:

\begin{equation}
    \mathcal{L}_{classed}=\frac{1}{N}\sum_{i}\sum_{j}^{|T^p|}f(T_{i, j}^{p})log(\text{Softmax}(\theta(h_{i}^{C}))),
\end{equation}
where the $\theta$ means the parameters of a MLP, and $f(T_{i, j}^{p})$ is the one-hot encoding of $j$-th parent category for the $i$-th sample. By classified on parent categories, we introduce a softer constraint that alleviates the challenge of fitting subcategories. This approach enables the model to capture the overall characteristics and shared features within the parent category, while still accommodating the specific details and variations within the subcategories. 
To obtain the final loss, we combine the two aforementioned losses. By summing these losses, we ensure that the model optimizes both the contrastive learning task and the aggregation of point cloud features under the guidance of structured vision and language information. 

\begin{table*}[]
    \caption{The results of zero-shot 3D classification on ModelNet40 and ScanObjectNN datasets. PointMLP + JM3D outperforms the previous state-of-the-art methods by a large margin in various evaluation settings, especially achieving a 12.3\% and 13.7\% improvement of the ``Medium'' and ``Hard'' mode on ModelNet40, which is the SOTA.}
    \centering
    \begin{tabular}{lcccccccc}
         \toprule
         \multirow{3}*{Model} & \multicolumn{6}{c}{ModelNet40} & \multicolumn{2}{c}{ScanObjectNN} \\
         
         \cmidrule(lr){2-7}\cmidrule(lr){8-9}
         
         ~ & \multicolumn{2}{c}{ All } & \multicolumn{2}{c}{Medium} & \multicolumn{2}{c}{Hard} & \multicolumn{2}{c}{All}
         \\
         \cmidrule(lr){2-3}\cmidrule(lr){4-5}\cmidrule(lr){6-7}\cmidrule(lr){8-9}
         ~  & top-1 & top-5 & top-1 & top-5 & top-1 & top-5 & top-1 & top-5
         \\
         \midrule
         PointCLIP~\cite{zhang2022pointclip}  & 20.2 & -- & 10.4 & --& 8.3 &-- & 15.4 & --\\
         PointMLP~\cite{marethinking} + CG3D~\cite{hegde2023clip} & 50.4 & -- & -- & -- & -- & -- & 25.0 & --\\
         PointTransformer~\cite{zhao2021point} + CG3D~\cite{hegde2023clip} & 50.6 & -- & -- & -- & -- & -- & 25.6 & --\\
         PointNet++(ssg)~\cite{qi2017pointnet++} + ULIP~\cite{xue2022ulip}& 55.7 & 75.7 & 35.6 & 64.4 & 33.7 & 55.8 & 45.6 & 73.8\\
         PointBERT~\cite{yu2022point} + ULIP~\cite{xue2022ulip}& 60.4 & \textbf{84.0} & 40.4 & 72.1 & 37.1 & 66.3 & 48.5 & 79.9\\
         PointMLP~\cite{marethinking} + ULIP~\cite{xue2022ulip} & 61.5 & 80.7 & 43.2 & 72.0 & 36.3 & 65.0 & 44.6 & 82.3\\
         \midrule
         PointNet++(ssg)~\cite{qi2017pointnet++} + JM3D~\cite{DBLP:conf/mm/WangTJSZMZLZLJ23} & 62.2 \textcolor{darkgreen}{\small ($\uparrow 6.5$)} & 79.3 & 47.6 \textcolor{darkgreen}{\small ($\uparrow 12.0$)} & 75.9 & 43.3 \textcolor{darkgreen}{\small ($\uparrow 9.6$)} & 74.7 & 46.0 \textcolor{darkgreen}{\small ($\uparrow 0.4$)} & 78.1\\
         
         PointBERT~\cite{yu2022point} + JM3D & 61.8 \textcolor{darkgreen}{\small ($\uparrow 1.4$)} & 81.7 & 52.9 \textcolor{darkgreen}{\small ($\uparrow 12.5$)} & 73.6 & 48.4 \textcolor{darkgreen}{\small ($\uparrow 11.3$)} & 71.1 & \textbf{48.9} \textcolor{darkgreen}{\small ($\uparrow 0.4$)} & 82.8\\
         
         PointMLP~\cite{marethinking} + JM3D & \textbf{65.8} \textcolor{darkgreen}{\small ($\uparrow 4.3$)} & 82.1 & \textbf{55.5} \textcolor{darkgreen}{\small ($\uparrow 12.3$)} & \textbf{77.1} & \textbf{55.0} \textcolor{darkgreen}{\small ($\uparrow 13.7$)} & \textbf{75.0} & 47.5 \textcolor{darkgreen}{\small ($\uparrow 2.9$)} & \textbf{83.3}\\

         \bottomrule
    \end{tabular}
    \label{tab:zero-shot-modelnet}
\end{table*}

\section{Integrating JM3D with Large Language Model}
Following the training on JM3D, the point cloud features gravitate towards linguistic features within a unified semantic space. This alignment suggests that the distributions of the point cloud features mirror their linguistic counterparts. With this foundation, we can harness Natural Language Processing (NLP) techniques to interpret point cloud data and address tasks demanding richer semantic understanding, such as crafting intricate 3D descriptions. The recent emergence of prompt tuning and SFT-based methods~\cite{liu2023visual, ouyang2022training} has streamlined diverse NLP tasks, showcasing formidable reasoning prowess and adaptability across a range of applications. Pioneering multi-modal research~\cite{zhu2023minigpt, touvron2023llama} has intertwined image data with language, leveraging Large Language Models (LLM) as foundational reasoning aids. Drawing inspiration from these advances, we augment JM3D with LLM, leading to our refined model: JM3D-LLM. This hybrid aims to probe the potentials of 3D models within a sophisticated semantic framework.

\subsection{Instruct Conversations for Point Querying}

Drawing from methodologies in recent studies~\cite{liu2023visual, zhu2023minigpt}, we first curated a comprehensive conversational dataset to fine-tune our model. Given that JM3D's pre-training data is primarily centered around individual object models rather than entire scenes, we sought point cloud data that would harmoniously complement this domain. Thus, we sourced data from Cap3D~\cite{luo2023scalable}, a database rich in detailed descriptions of various objects.

For our instruction-based tasks, we employed a captioning strategy. The objective is to empower the LLM to generate in-depth descriptions of 3D models, effectively maximizing the semantic depth extracted from point cloud representations. In doing so, the LLM is enriched with a more granular understanding of the 3D spatial information. For our implementation, we adhered to the criteria outlined in Tab.~\ref{tab:instruct}, ensuring the diverse and consistent framing of instructional data.

\subsection{JM3D-LLM Architecture}

Within our JM3D-LLM architecture as shown in Fig.~\ref{fig4}, a Multi-layer Perceptron (MLP) is deployed to align the feature dimensions of the point cloud with those of the language tokens. Subsequently, these point cloud features are seamlessly integrated as specialized tokens into the LLM's input stream.

For more details, given a paragraph from a conversation, a tokenizer, denoted as \(f_t\), converts all tokens into their corresponding IDs using the vocabulary \(f_v\). In parallel, the pre-trained point cloud module, represented by \(f_{c_{pretrain}}\), gleans a sequence of point cloud features, symbolized as \(T_c = \{c_1, c_2, \dots, c_n\} \in \mathbb{R}^{n \times d_1}\), with \(n\) signifying the token count of the point features.

These point cloud features undergo a transformation, as represented by:
\begin{equation}
    T^{'}_c = \text{MLP}(T_c) \rightarrow D_{llm}(t; f_v, f_t),
\end{equation}
where \(D_{llm}(t; f_v, f_t)\) denotes the token distribution within the LLM framework. The adjusted tokens, \(T^{'}_c\), are then positioned at placeholders marked by \(\langle \text{point} \rangle\). The LLM, represented as \(f_{llm}\), subsequently processes the remaining language tokens to produce a sequence of linguistic features, \(T_l = \{l_1, l_2, \dots, l_m\} \in \mathbb{R}^{m \times d}\). The aggregate set of tokens can be represented as \( [T^{'}_c, T_l] \in \mathbb{R}^{(n+m) \times d_1}\).

Post integration of these tokens into \(f_{llm}\), the LLM iteratively derives the associated hidden states, represented by \(z_i\). These states are then projected onto the vocabulary space \(f_v\) to yield \(\widetilde{z_i}\). Leveraging a softmax operation, the resultant distribution is created over the vocabulary, leading to the decoding of the word \(\widetilde{w}\) with the maximal probability, as illustrated by:
\begin{equation}
    \widetilde{w}_i = f^{-1}_v(\mathop{\arg \max}\limits_{w \in \widetilde{z}_i} \text{Softmax}(z_i)).
\end{equation}

\begin{figure*}[t]
\centering
\includegraphics[width=1.0\textwidth]{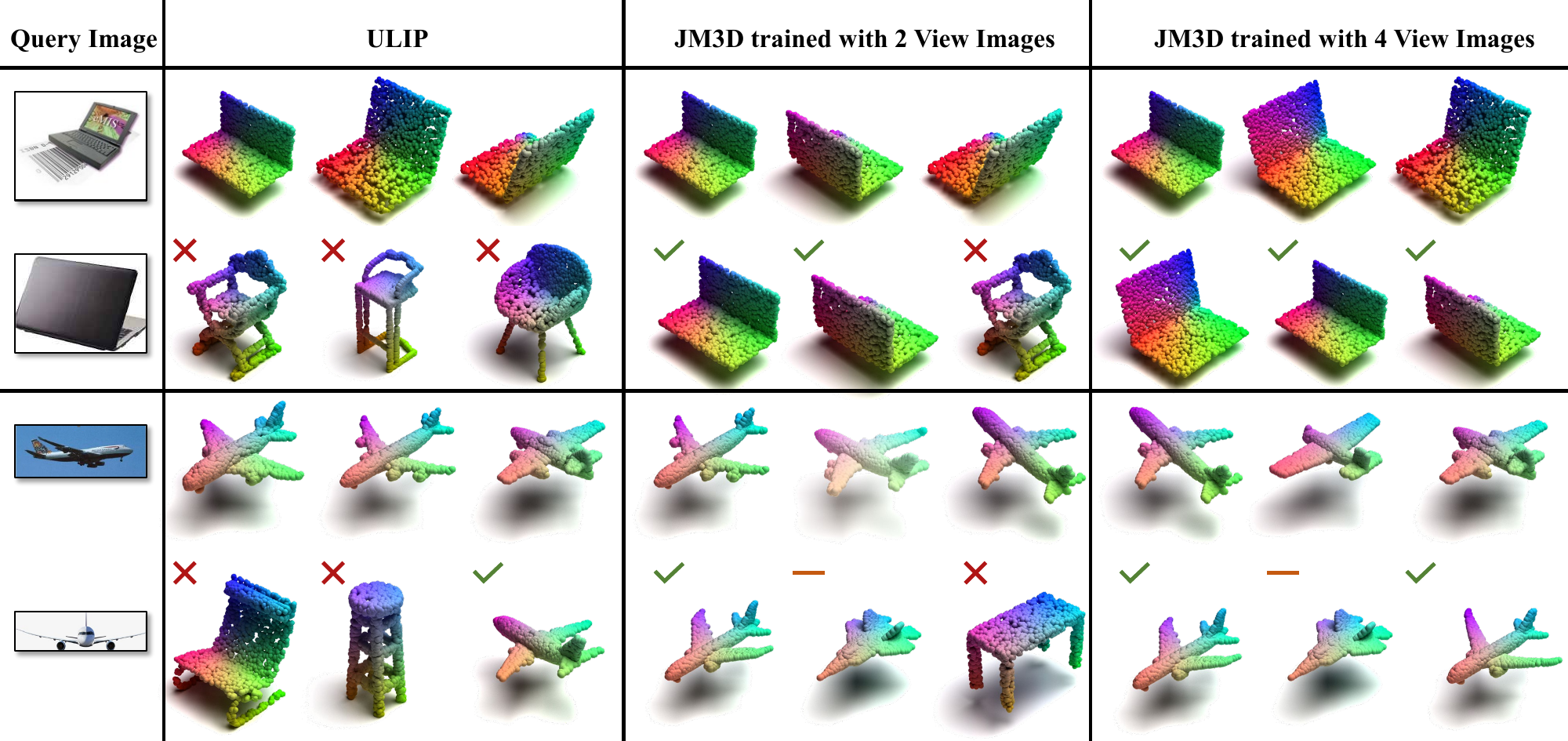}
\caption{The qualitative results of the real image to point cloud retrieval. Giving an image, We show the top-3 point cloud retrieval results from ModelNet40. All models perform well on the simple samples (the 1st row and the 3rd row). However, when it comes to the challenging samples (the 2nd row and the 4th row), JM3D demonstrates a more accurate retrieval ability compared to the previous state-of-the-art (ULIP). The JM3D trained with 4 view images shows better performance compared with the 2 view images, benefiting the more solid bias of vision modality.}
\label{fig3}
\end{figure*}

\section{Experiment}
\subsection{Datasets}
\subsubsection{Pretrain Dataset}
We use \textbf{ShapeNet55}~\cite{chang2015shapenet} as the pre-training dataset, which is the publicly-available subset of ShapeNet. ShapeNet consists of 52.5K CAD models with multiple texture maps and corresponding category annotations. The annotations have a total of 55 basic categories and 205 fine-grained subcategories, with a small number of models missing subcategories. During training, we randomly sample different numbers of points from the CAD models to adapt different networks.




\subsubsection{Downstream Datasets}
Our investigations in downstream tasks are predominantly anchored on two datasets, offering diverse perspectives on 3D object categorization.

\textbf{ModelNet40}~\cite{wu20153d}: 
Originating from synthetic 3D CAD models, this dataset amasses 9,843 training samples and 2,468 testing samples, distributed over 40 distinct categories. When testing, we adhere to the protocol outlined in~\cite{marethinking}, which entails downsampling the point cloud data to a resolution of 1024 points.

\textbf{ScanObjectNN}~\cite{uy2019revisiting}: 
In contrast to ModelNet40, ScanObjectNN offers a collection of 3D objects that have been scanned directly from real-world scenarios. This dataset houses 2,902 samples, spread across 15 categories. Intriguingly, it presents two variations, distinguished by the presence or absence of background noise: \emph{OBJ\_ONLY} and \emph{OBJ\_BJ}. The former encapsulates a pristine mesh, devoid of any background interference, while the latter retains background noise. In our studies, we utilize the pre-processed data as recommended by ULIP~\cite{uy2019revisiting}, which has been further refined by~\cite{yu2022point}. This data is both normalized and downsampled to 1024 points for uniformity.

\begin{table*}[]
    \caption{The ablation study of CIS. Multi-views will lead to point cloud features being biased towards vision modality, which decreases the performance on zero-shot 3D classification. However, CIS effectively improves the performance with the embeddings and within-view sample.}
    \centering
    \begin{tabular}{ccc|cccccccc}
         \toprule
         \multirow{3}*{Views} & \multirow{3}*{Sample} & \multirow{3}*{Embedding}  & \multicolumn{6}{c}{ModelNet40} & \multicolumn{2}{c}{ScanObjectNN} \\
         
         \cmidrule(lr){4-9}\cmidrule(lr){10-11}
         
         ~ & ~ & ~ & \multicolumn{2}{c}{ All } & \multicolumn{2}{c}{Medium} & \multicolumn{2}{c}{Hard} & \multicolumn{2}{c}{All}
         \\
         \cmidrule(lr){4-5}\cmidrule(lr){6-7}\cmidrule(lr){8-9}\cmidrule(lr){10-11}
         ~ & ~ & ~ & top-1 & top-5 & top-1 & top-5 & top-1 & top-5 & top-1 & top-5 \\
         \midrule
         1 & random & \ding{55} & 60.0 & 79.4 & 43.2 & 72.0 & 36.3 & 65.0 & 44.6 & 82.3\\
         4 & random & \ding{55} & 56.5 & 76.6 & 42.3 & 70.4 & 38.2 & 66.3 & 44.5 & 75.3\\
         4 & random & \ding{51} & 58.8 & 79.9 & 47.6 & \textbf{75.1} & 47.5 & 71.2 & 44.9 & 81.2\\
         4 & within-view & \ding{51} & 60.7 & 80.2 & 48.7 & 74.7 & 50.0 & 71.2 & \textbf{45.4} & \textbf{82.8}\\
         2 & within-view & \ding{51} & \textbf{61.2} & \textbf{80.7} & \textbf{49.5} & 74.5 & \textbf{51.4} & \textbf{71.7} & \textbf{45.4} & 82.2\\

         \bottomrule
    \end{tabular}
    \label{tab:ablation-image}
\end{table*}

\begin{table*}[]
    \caption{The ablation study of HTT. Results show that structured text is more effective than fine-grained text. Even if the language modality is enhanced, the independent alignment method still makes the improvements unstable.}
    \centering
    \begin{tabular}{cc|cccccccc}
         \toprule
         \multirow{3}*{Contrastive Text} & \multirow{3}*{Classed Text}  & \multicolumn{6}{c}{ModelNet40} & \multicolumn{2}{c}{ScanObjectNN} \\
         
         \cmidrule(lr){3-8}\cmidrule(lr){9-10}
         
         ~ & ~ & \multicolumn{2}{c}{ All } & \multicolumn{2}{c}{Medium} & \multicolumn{2}{c}{Hard} & \multicolumn{2}{c}{All}
         \\
         \cmidrule(lr){3-4}\cmidrule(lr){5-6}\cmidrule(lr){7-8}\cmidrule(lr){9-10}
         ~ & ~ & top-1 & top-5 & top-1 & top-5 & top-1 & top-5 & top-1 & top-5 \\
         \midrule
         Parent category & \ding{55} & 61.2& 80.7 & 49.5 & 74.5 & \textbf{51.4} & 71.7 & 45.4 & \textbf{82.2}\\
         Subcategory & \ding{55} & 61.8 & 81.0 & 48.5 & \textbf{78.6} & 43.5 & \textbf{75.2} & 46.4 & 79.5\\
         Subcategory & Parent category & \textbf{63.1} & \textbf{81.4} & \textbf{49.7} & 76.7 & 50.4 & 74.4 & \textbf{46.9} & 80.4\\

         \bottomrule
    \end{tabular}
    \label{tab:ablation-text}
\end{table*}

\subsubsection{Instruct Dataset}
We source our instructional fine-tuning conversation data from Cap3D~\cite{luo2023scalable}, which is a rich repository of point cloud samples curated from Objverse~\cite{deitke2023objaverse}, accompanied by detailed descriptions. Specifically, Cap3D boasts 660k point cloud data instances, each paired with a description. Notably, 40k of these descriptions undergo manual annotation for enhanced precision. For instructive purposes, we adopt the methodology outlined in~\cite{touvron2023llama}, devising 11 tailor-made templates tailored to the nuances of 3D modality.

\subsection{3D Backbone Networks}
To ascertain the efficacy of our novel JM3D framework, we orchestrate experiments on ModelNet40 and ScanObjectNN, employing a variety of 3D backbone networks, namely:

\textbf{PointNet++}~\cite{qi2017pointnet++}: 
Serving as the successor to PointNet~\cite{qi2017pointnet}, PointNet++ harnesses an encoder-decoder architecture to delve deep into hierarchical features of point sets. Its encoder consists of a plethora of set abstraction modules interspersed with farthest point sampling techniques, the latter aiming to condense the scale of point sets. The decoder, meanwhile, channels the outputs of the encoder's multiple layers to a range of head networks, ensuring adaptability to a spectrum of tasks.

\textbf{PointMLP}~\cite{marethinking}: 
In contrast, PointMLP is a streamlined network, integrating residual MLP modules for enhanced feature extraction, all while sidestepping intricate architectures. It innovatively incorporates a Geometric Affine Module, designed to transition points to a standard distribution. Two distinct MLP blocks then undertake the task of gleaning both representational and positional information.

\textbf{PointBert}~\cite{yu2022point}: 
A model rooted in the transformer paradigm, PointBert integrates self-supervised learning into point cloud representation. Its prowess lies in reconstructing masked point clouds, a feature that has yielded stellar outcomes on unlabeled point cloud datasets. In essence, PointBert strategically obscures select points, subsequently refining features through the act of reconstructing these occluded points.

\subsection{Implementation Details}

\subsubsection{Pre-training}
We uniformly sample the point cloud into 1024, 2048, and 8192 points to align with the recommended settings of various backbones. The preprocessing of rendered images and texts adheres to the requirements of the pre-trained image-text encoder. Favoring performance, we opt for SLIP~\cite{mu2022slip} over the conventional CLIP model. In our setup, both the image and text encoders remain static, mirroring the behavior of ULIP. Throughout training, only the parameters of the point cloud backbone undergo adjustments. JM3D undergoes training for 250 epochs, with batches of 128 and a learning rate set at $1e-3$. For optimization, we employ AdamW alongside the Cosine LR schedule.

\subsubsection{Zero-shot 3D Classification}
JM3D evaluates the distance between point cloud features and text features from new datasets, selecting the category with the shortest distance. The preprocessing steps for both text and point cloud mirror those of pre-training, eliminating the need for a fine-tuning phase. Our zero-shot evaluations target both ModelNet40 and ScanObjectNN. While ModelNet40, with its synthetic nature and unseen categories, serves as a benchmark for evaluating the alignment capabilities of multi-modal features, ScanObjectNN, populated with real-world scanned data, tests the resilience of 3D pre-trained models.

\begin{table}[]
    \caption{The ablation study for JMA. Independent alignment wastes the rich semantics brought by SMO, while JMA achieves significant improvements on all settings of different datasets.}
    \centering
    \resizebox{0.98\columnwidth}{!}{
    \begin{tabular}{ccc|cccccccc}
         \toprule
         \multirow{3}*{CIS} & \multirow{3}*{HTT} & \multirow{3}*{JMA}  & \multicolumn{6}{c}{ModelNet40} & \multicolumn{2}{c}{ScanObjectNN} \\
         
         \cmidrule(lr){4-9}\cmidrule(lr){10-11}
         
         ~ & ~ & ~ & \multicolumn{2}{c}{ All } & \multicolumn{2}{c}{Medium} & \multicolumn{2}{c}{Hard} & \multicolumn{2}{c}{All}
         \\
         \cmidrule(lr){4-5}\cmidrule(lr){6-7}\cmidrule(lr){8-9}\cmidrule(lr){10-11}
         ~ & ~ & ~ & top-1 & top-5 & top-1 & top-5 & top-1 & top-5 & top-1 & top-5 \\
         \midrule
         \ding{55} & \ding{55} & \ding{55} & 60.0 & 79.4 & 43.2 & 72.0 & 36.3 & 65.0 & 44.6 & 82.3\\
         \ding{51} & \ding{51} & \ding{55} & 63.1 & 81.4 & 49.7 & 76.7 & 50.4 & 74.4 & 46.9 & 80.4\\
         \ding{51} & \ding{51} & \ding{51} & \textbf{65.8} & \textbf{82.1} & \textbf{55.5} & \textbf{77.1} & \textbf{55.0} & \textbf{75.0} & \textbf{47.5} & \textbf{83.3}\\

         \bottomrule
    \end{tabular}}
    \label{tab:ablation-joint}
\end{table}

\subsubsection{JM3D-LLM}
For JM3D-LLM, we integrate the 7B Vicuna~\cite{chiang2023vicuna} as its foundation for LLM and employ PointMLP~\cite{marethinking}, known for its superior performance, as our point cloud encoder. A linear layer bridges the point cloud dimensions with the language domain. To enhance the point cloud feature set, we exclude the terminal classification layer of PointMLP, capturing 64 tokens from the point cloud for embedding into the query. The primary learning rate stands at $2e-3$, but PointMLP operates at a diminished $2e-5$, with the LLM remaining unaltered. We carry out all experiments on three 40G A100s.

\subsection{Zero-shot 3D Classification}

JM3D facilitates zero-shot 3D recognition by calculating the similarity between point cloud and textual features. In this segment, we exhibit results underscoring JM3D's aptitude for cross-modal understanding through zero-shot evaluations on two distinct datasets. Our primary evaluation metric is the top-1 accuracy, reflecting the efficacy of the feature alignment in real-world applications.

\subsubsection{Evaluation Sets}
To maintain consistency with prior research~\cite{xue2022ulip, zhang2022pointclip, hegde2023clip}, we perform experiments using a single model on the ModelNet40 and ScanObjectNN datasets. The results for all test samples are summarized in the ``All'' column of Tab.~\ref{tab:zero-shot-modelnet}. Additionally, to address potential confounding factors arising from the presence of similar classes in both the ModelNet40 and pre-trained datasets, which can significantly impact the fairness of zero-shot evaluation, we have created two distinct subsets within ModelNet40: a "Medium" set and a ``Hard'' set. 
The ``Medium'' set excludes classes that are common to both ModelNet40 and ShapeNet55, while the ``Hard'' set goes a step further by eliminating semantically similar classes. For instance, the class ``stool'' is excluded from the ``Hard'' set as it already exists in the ShapeNet55 dataset. This rigorous selection process ensures that all categories within the ``Hard'' set remain uncompromised by any potential information leakage. 
For each of these subsets, we employ top-1 accuracy and top-5 accuracy as our evaluation metrics. Top-1 accuracy assesses the model's capability to correctly identify the nearest text representation to the point cloud feature. We believe that the top-1 metric provides a more intuitive demonstration of the model's zero-shot performance.

\subsubsection{Experiment Results}
In Tab.~\ref{tab:zero-shot-modelnet}, we provide a comprehensive overview of our zero-shot evaluations conducted on the ModelNet40 and ScanObjectNN datasets. Remarkably, our JM3D model surpasses the former state-of-the-art ULIP~\cite{xue2022ulip} across all 3D backbones, exhibiting superior top-1 accuracy. Specifically, when comparing JM3D+PointMLP to ULIP, we observe a substantial improvement in top-1 accuracy: a 4.3\% increase on the ``All'' set, a 12.3\% enhancement on the ``Medium'' set, and an impressive 13.7\% boost on the ``Hard'' set. These results emphasize the clear superiority of JM3D in the context of zero-shot learning. 
Furthermore, our evaluations extend to the ScanObjectNN dataset, where JM3D + PointMLP outperforms ULIP by a notable 2.9\% margin in terms of top-1 accuracy. In summary, our findings represent significant advancements over the previous state-of-the-art method, ULIP~\cite{xue2022ulip}. They underscore JM3D's robust generalizability and its ability to deliver enhanced performance in real-world 3D scanning scenarios.

\begin{table}[]
    \small
    \centering
    \caption{The semantic segmentation on S3DIS.}
    \resizebox{0.98\columnwidth}{!}{
    \begin{tabular}{lcc}
    \toprule
         Model& Overall Acc & Class avg IoU \\
         \midrule
         PointNet++ (ssg) \cite{qi2017pointnet++} &  83.0 & 53.5 \\
         PointNet++ (ssg) \cite{qi2017pointnet++} + JM3D &  \textbf{83.6} & 
 \textbf{57.1}\\
         \bottomrule
    \end{tabular}
    }
    \label{tab:seg-semantic}
\end{table}

\begin{table}[]
    \small
    \centering
    \caption{The Part segmentation on ShapeNet.}
    \resizebox{0.98\columnwidth}{!}{
    \begin{tabular}{lcc}
    \toprule
         Model& Instance avg IoU & Class avg IoU \\
         \midrule
         3D-GCN~\cite{lin2020convolution} & 85.1 & 82.1 \\
         SageMix \cite{lee2022sagemix} &  85.4 & - \\
         SFCNN~\cite{rao2019spherical} & 85.4 & \textbf{82.7} \\
         DGCNN~\cite{sun2022self} & \textbf{85.5} & - \\
         \midrule
         PointNet++ (ssg) \cite{qi2017pointnet++} &  84.9 & 81.8 \\
         PointNet++ (ssg) \cite{qi2017pointnet++} + JM3D &  \textbf{85.5} & 82.1\\
         \bottomrule
    \end{tabular}}
    \label{tab:seg-part}
\end{table}

\subsection{3D Classification Fine-tuning}
To demonstrate the capabilities of JM3D, we perform fine-tuning experiments on ScanObjectNN using one of the state-of-the-art frameworks, PointMLP~\cite{marethinking}.

During the fine-tuning process, we train only the 3D encoder of JM3D on the \emph{PB\_T50\_RS} subset of ScanObjectNN. We choose \emph{PB\_T50\_RS} due to its challenging nature: it consists of real-world scanned objects accompanied by background noise. The PointMLP fine-tuning uses a learning rate of 0.03 and a weight decay of 3e-4 over 350 epochs. We initiate the process with the pre-trained parameters and maintain the original model's conditions throughout. Consistent with standard conventions in the research community, we use OA (Overall Accuracy) and mAcc (Class Average Accuracy) as evaluation metrics.

Tab.~\ref{tab:fintune-scan} illustrates that JM3D substantially boosts the performance of the baseline model. Specifically, JM3D augments PointMLP's OA by 3.5\% and mAcc by 4.0\%. Moreover, PointMLP enhanced with JM3D surpasses the previous state-of-the-art, RepSurf-U ($2\times$), by a margin of 3.2\%. Employing a voting strategy, PointMLP$^{*}$ combined with JM3D establishes a new benchmark. Indeed, when compared to the training approach used in ULIP, JM3D clearly demonstrates its ability to enhance existing models without requiring custom-designed architectures.

\begin{table}[]
    \caption{3D classification results on ScanObjectNN. We follow the default settings of the original method to train on the hardest set. JM3D achieves a significant improvement compared to the previous method, helping to improve the original backbone by 3.5\%.}
    \centering
    \resizebox{\columnwidth}{!}{
    \begin{tabular}{lcc}
    \toprule
         Model& Overall Acc & Class-mean Acc \\
         \midrule
         PointNet \cite{qi2017pointnet} &  68.2 & 63.4 \\
         PointNet++ \cite{qi2017pointnet++} &  77.9 & 75.4 \\
         DGCNN \cite{wu2018dgcnn} &  78.1 & 73.6 \\
         MVTN \cite{hamdi2021mvtn} &  82.8 &  --\\
         PointBERT \cite{yu2022point} &  83.1 &  --\\
         RepSurf-U \cite{ran2022surface} & 84.6 &  --\\
         PointMAE \cite{pang2022masked} & 85.2 &  --\\
         RepSurf-U (2x) \cite{ran2022surface} &  86.0 &  --\\

         \midrule
         PointMLP \cite{marethinking} &  85.7 & 84.4 \\
         PointMLP+ ULIP &  88.8 & 87.8 \\
         PointMLP + JM3D & 89.2 & 88.4 \\
         PointMLP$^{*}$ + JM3D & \textbf{89.5} & \textbf{88.7} \\
         \bottomrule
    \end{tabular}}
    \label{tab:fintune-scan}
\end{table}

\begin{table*}[!h]
\centering
\caption{Qualitative comparisons with ULIP-based LLM and our JM3D-LLM on our benchmark.}

\resizebox{0.9\textwidth}{!}{
\begin{tabular}{@{}l p{0.4\linewidth} p{0.4\linewidth}}
\toprule

Samples 1,2 & 
  \begin{minipage}{\linewidth}
    \hspace{0.8em}
    \includegraphics[width=0.28\linewidth]{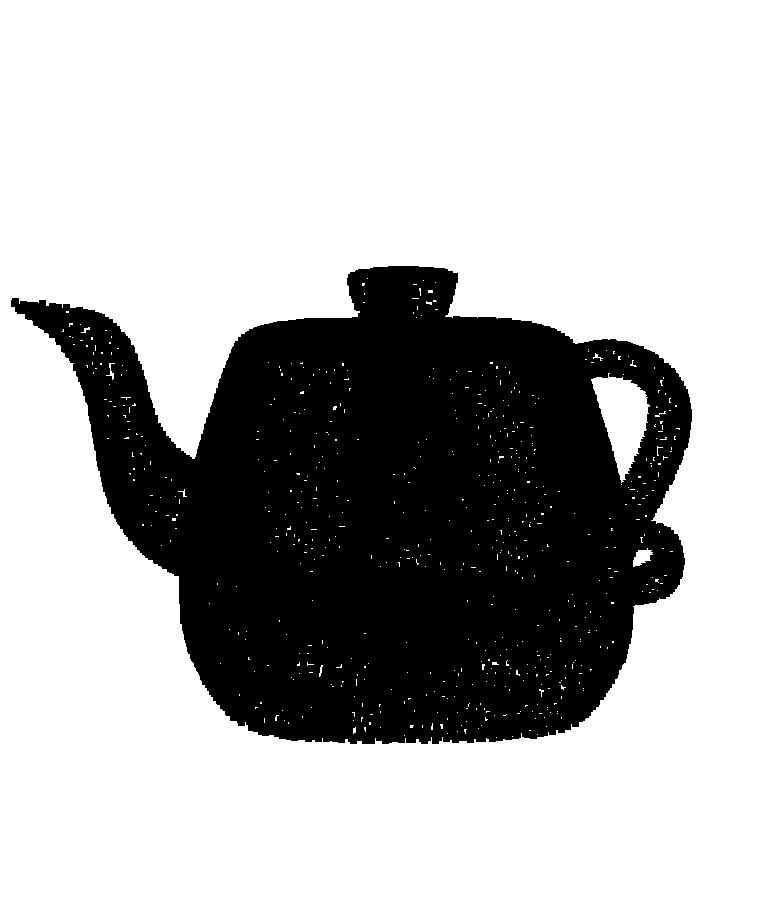}
    \hspace{4.7em}
    \includegraphics[width=0.28\linewidth]{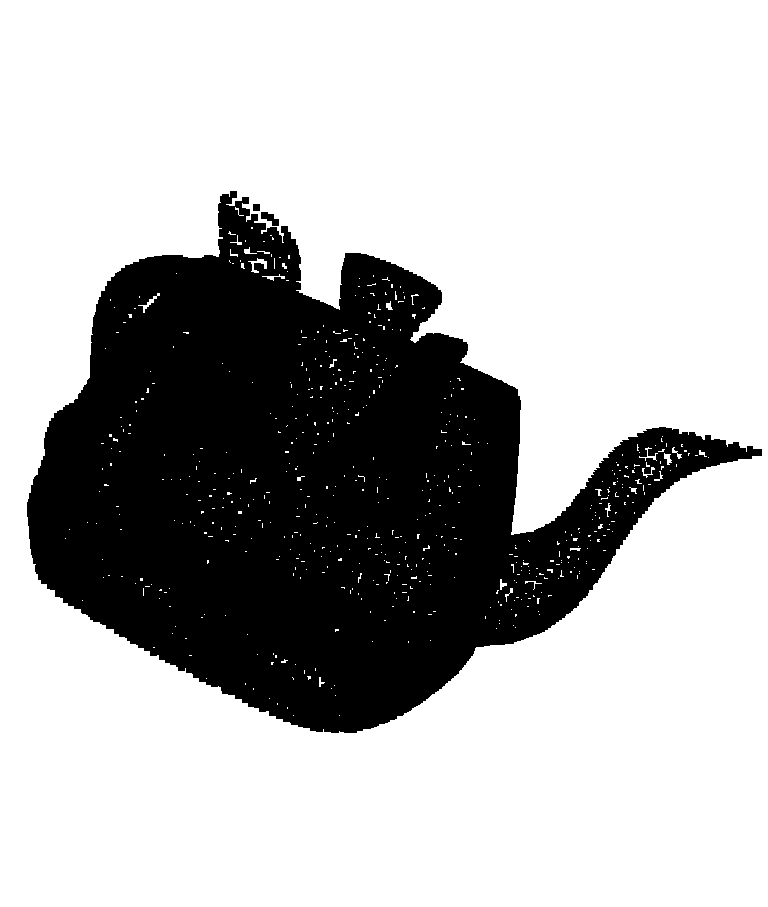}
  \end{minipage}
& 
  \begin{minipage}{\linewidth}
    \hspace{2.8em}
    \includegraphics[width=0.28\linewidth]{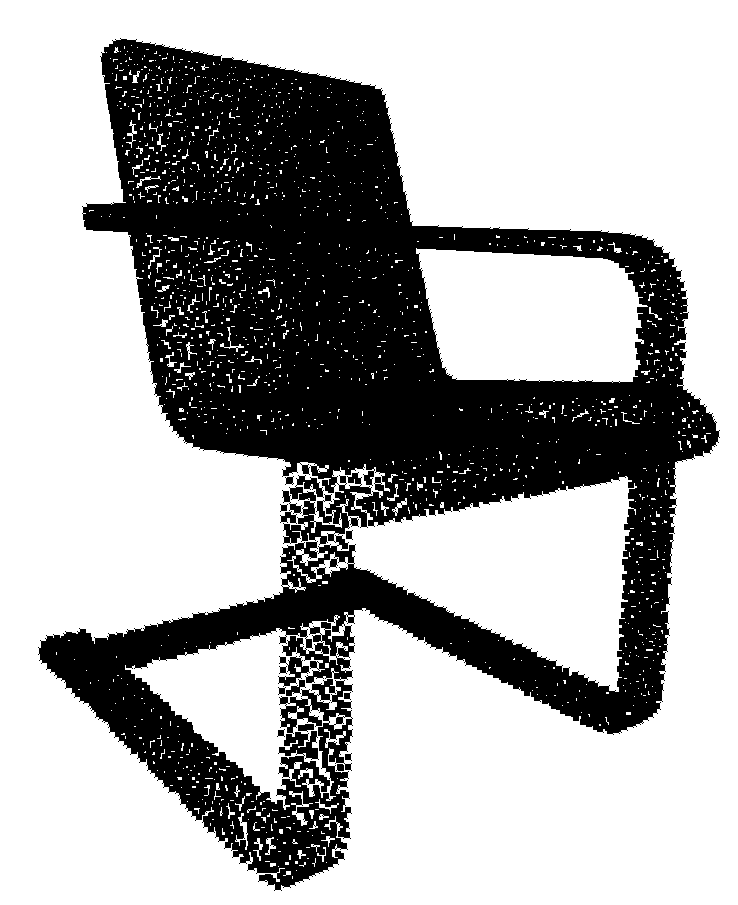}
    \hspace{3.7em}
    \includegraphics[width=0.28\linewidth]{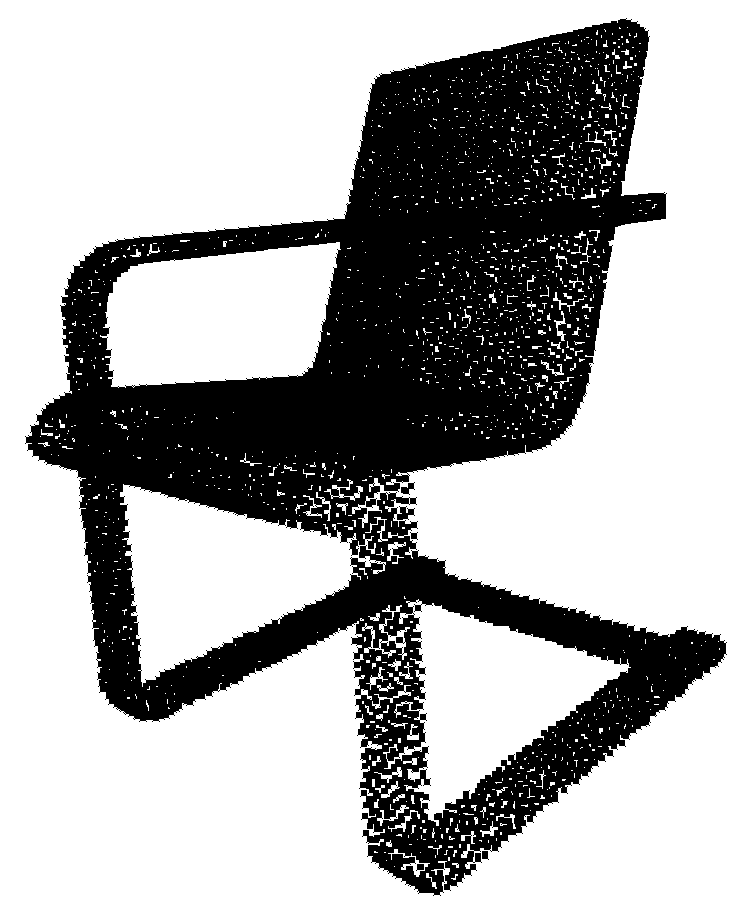}
  \end{minipage}
\\ \midrule
Uid & 36a8e79cf9c64148b356ea7500670b86 & f2e2daaaf78d49dc848240f7ff7562f3 \\
Ground Truth & A yellow horned teapot. & A modern green chair with a wooden frame and upholstered seat. \\ 
Category & Teapot & Chair \\
\midrule

User & What is this? & What is this? \\
ULIP-based LLM & A small teapot. & A chair with a wooden base. \\ 
JM3D-LLM & A small teal teapot. & A wooden chair with a seat and backrest. \\ 
\bottomrule

Samples 3,4 & 
  \begin{minipage}{\linewidth}
    \hspace{1.8em}
    \includegraphics[width=0.30\linewidth]{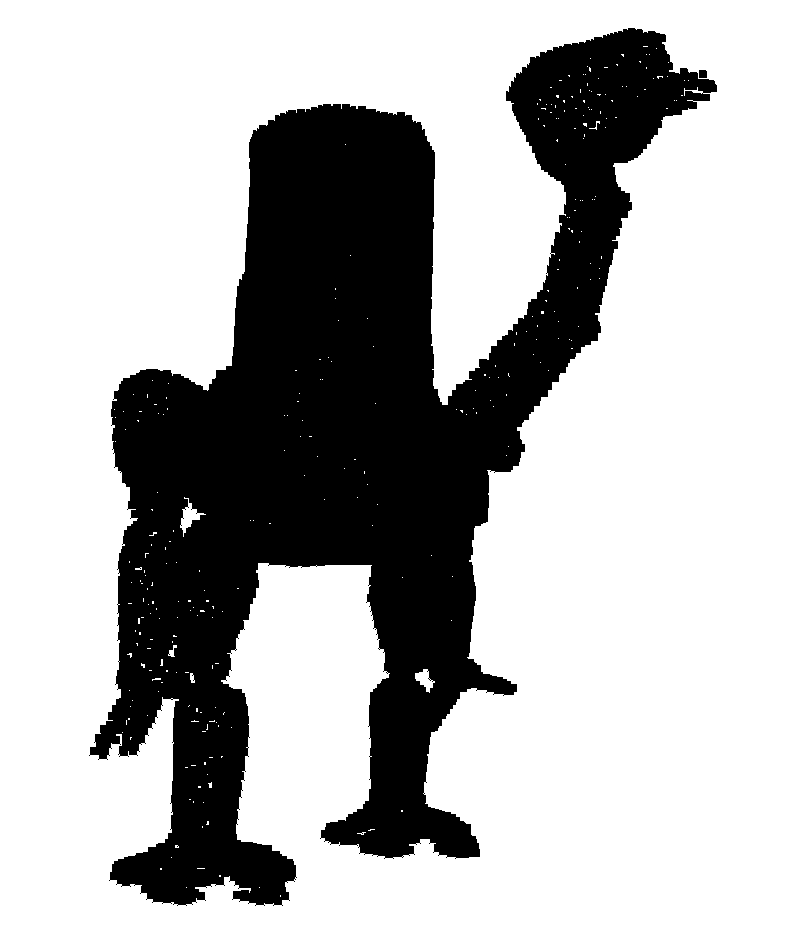}
    \hspace{3.0em}
    \includegraphics[width=0.30\linewidth]{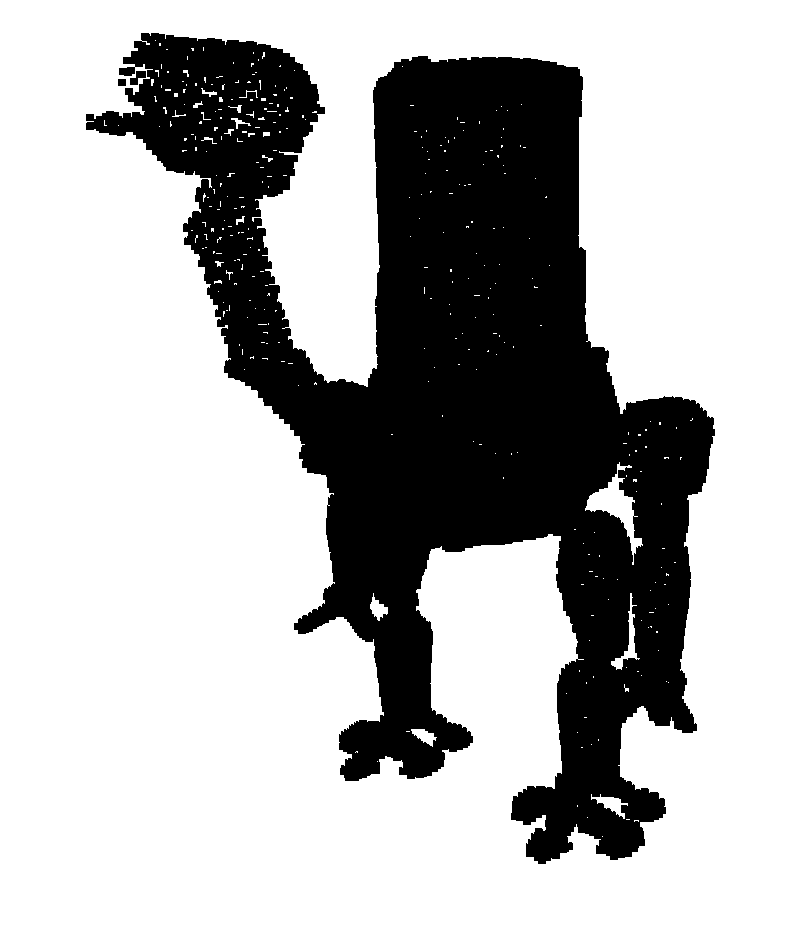}
  \end{minipage}
& 
  \begin{minipage}{\linewidth}
    \includegraphics[width=0.48\linewidth]{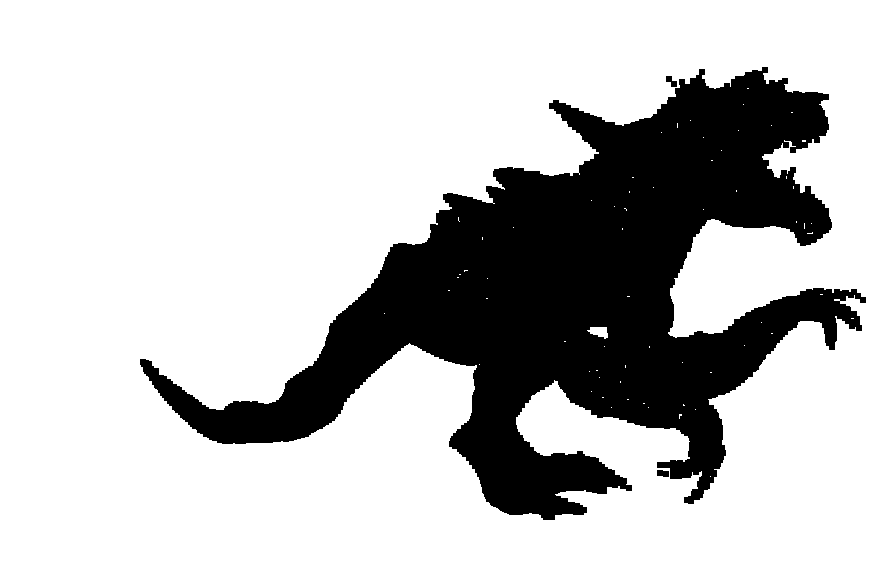}
    \hspace{2.7em}
    \includegraphics[width=0.48\linewidth]{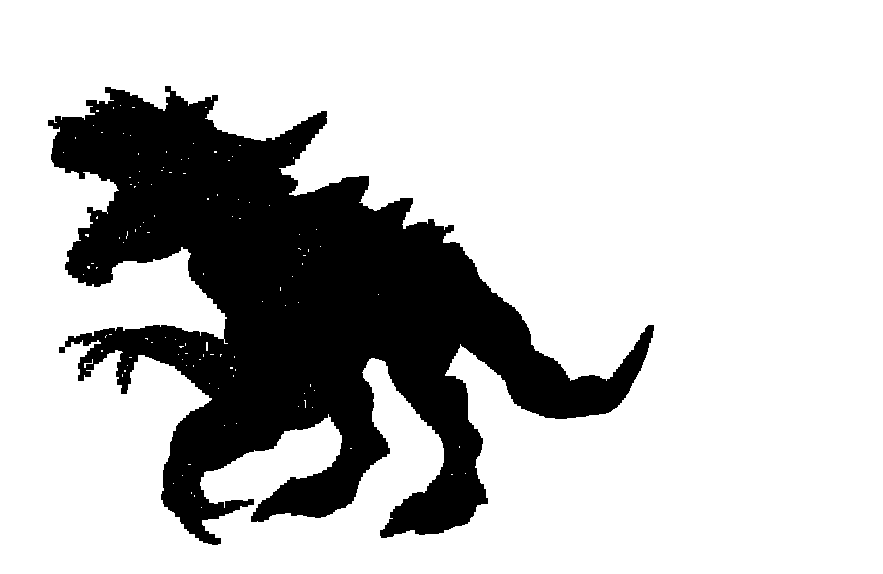}
  \end{minipage} \\ \midrule
Uid & 7e87939fb2e444658ea3f9cf33905e4d & e85ebb729b02402bbe3b917e1196f8d3 \\
Ground Truth & A 3D model of a red and green robot with a long neck holding a gun. & A 3D model of a black dragon-like monster with sharp claws and spikes. \\ 
Category & Robot & Dragon \\

\midrule

User & What is this? & What is this? \\
ULIP-based LLM & A cartoon character in a white outfit holding a sword. & A white and black spaceship with a sword. \\ 
JM3D-LLM & A white robot with arms and legs. & A white dragon-like creature with a long tail, resembling a bird and a lizard. \\ 
\bottomrule

Samples 5,6 & 
  \begin{minipage}{\linewidth}
    \includegraphics[width=0.36\linewidth]{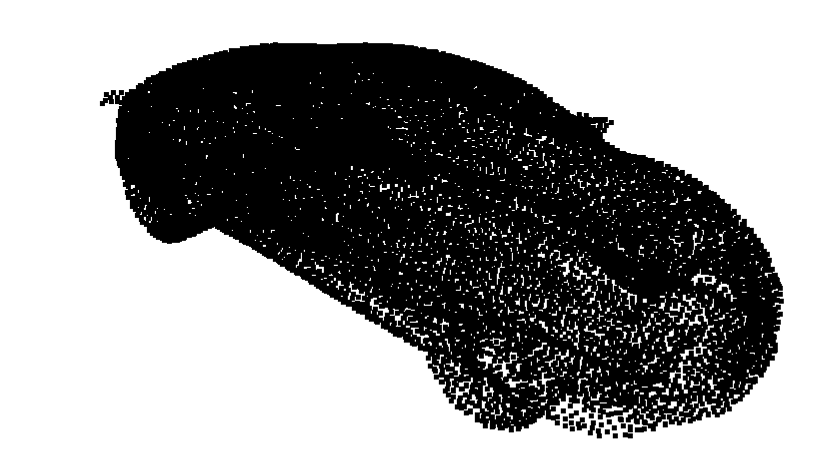}
    \hspace{3.7em}
    \includegraphics[width=0.36\linewidth]{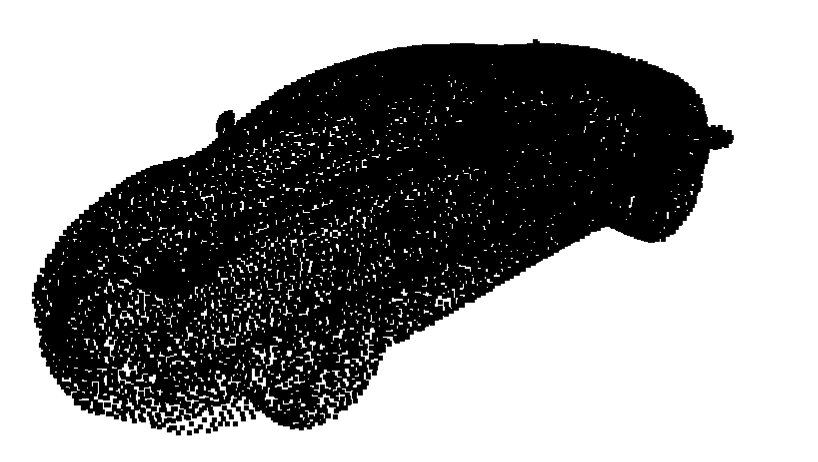}
  \end{minipage}
& 
  \begin{minipage}{\linewidth}
    \includegraphics[width=0.48\linewidth]{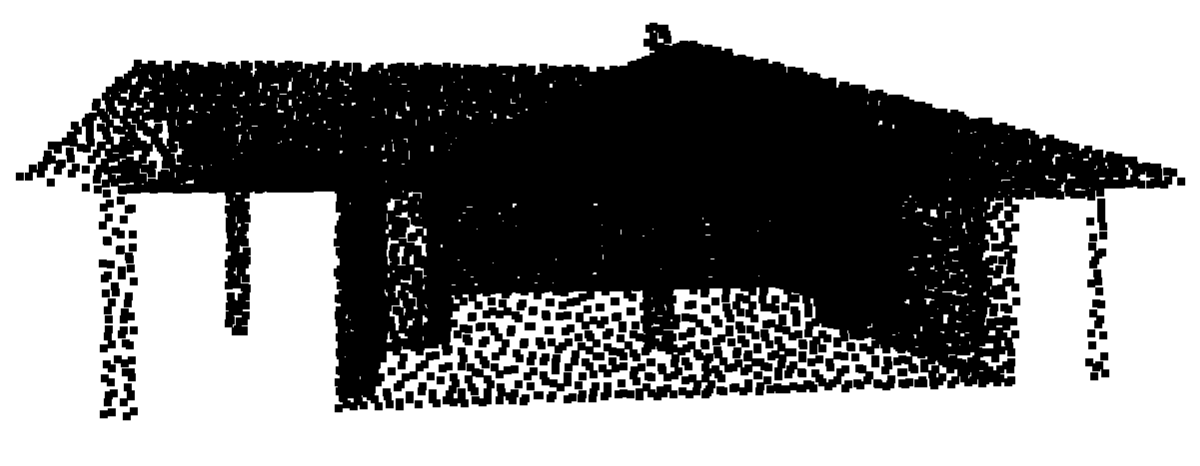}
    \includegraphics[width=0.48\linewidth]{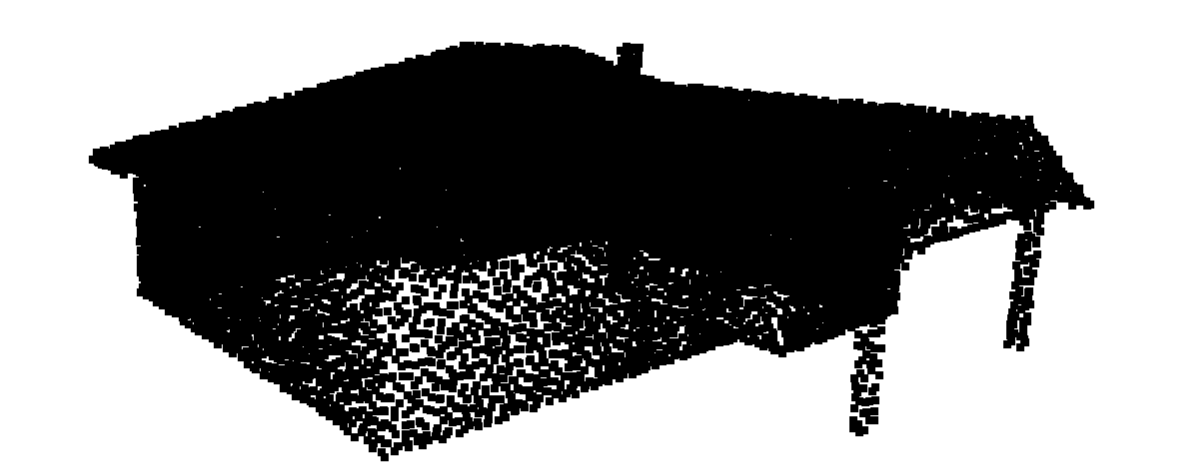}
  \end{minipage} \\ \midrule
Uid & d10f75f74355452f8e417cc36f253841 & 301a20db049b47e19c7d56a5ba5f504f \\
Ground Truth & Red Audi R8 3D model sports car. & 3D model of a house with a roof and wooden frame. \\ 
Category & Car & House \\
\midrule
User & What is this? & What is this? \\
ULIP-based LLM & A white sports car. & A house with a roof structure. \\ 
JM3D-LLM & A sports car with a futuristic design, resembling Audi R8 V10 Plus. & A house with a roof, featuring a roof structure and ceiling. \\ 
\bottomrule

Samples 7,8 & 
  \begin{minipage}{\linewidth}
    \hspace{1.8em}
    \includegraphics[width=0.24\linewidth]{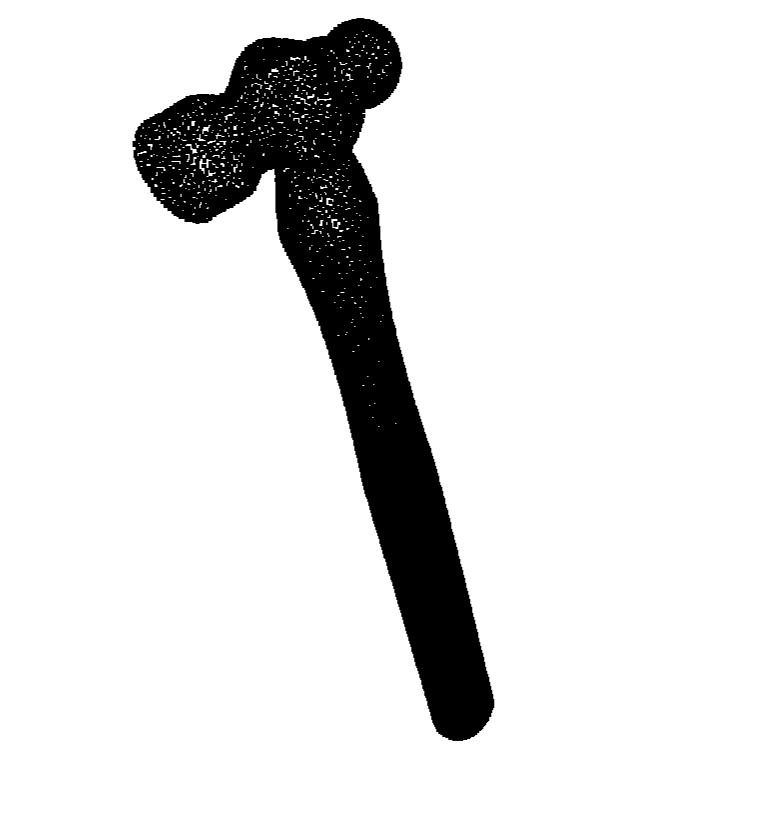}
    \hspace{4.8em}
    \includegraphics[width=0.24\linewidth]{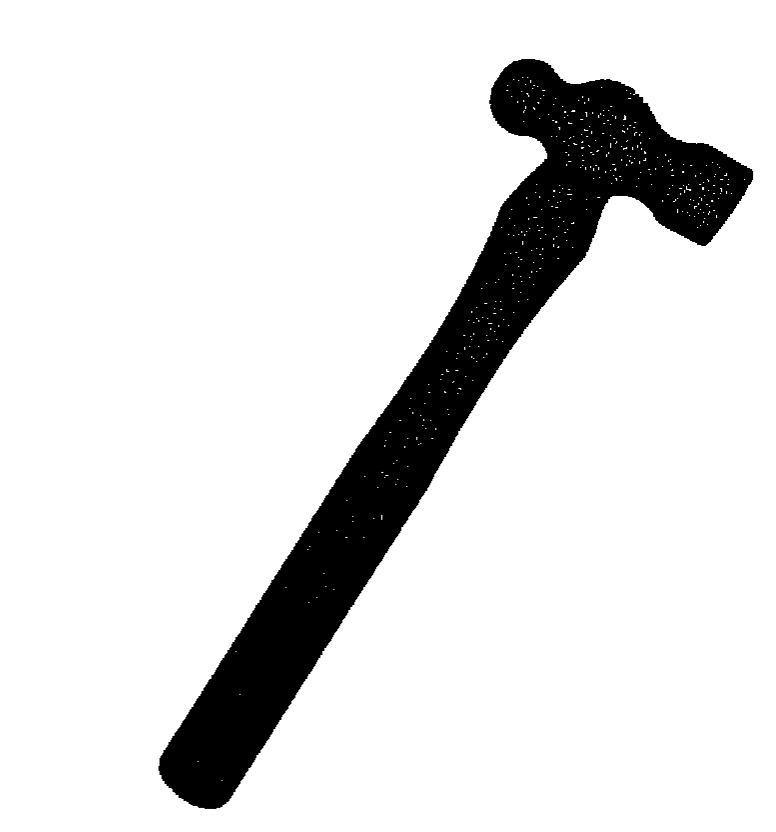}
  \end{minipage}
& 
  \begin{minipage}{\linewidth}
    \hspace{2.8em}
    \includegraphics[width=0.24\linewidth]{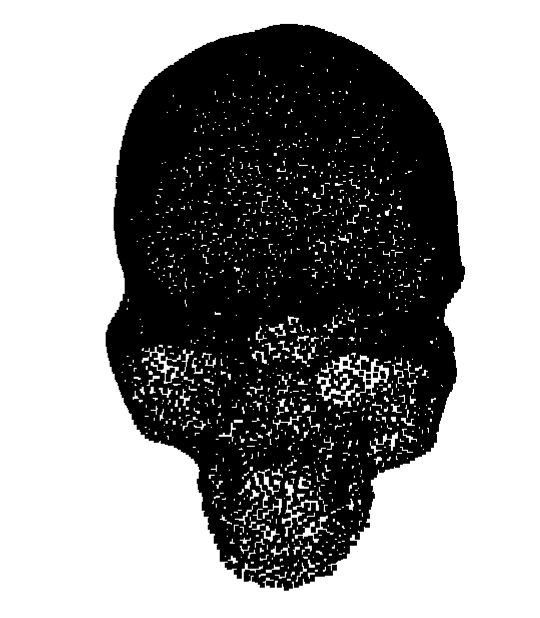}
    \hspace{4.7em}
    \includegraphics[width=0.24\linewidth]{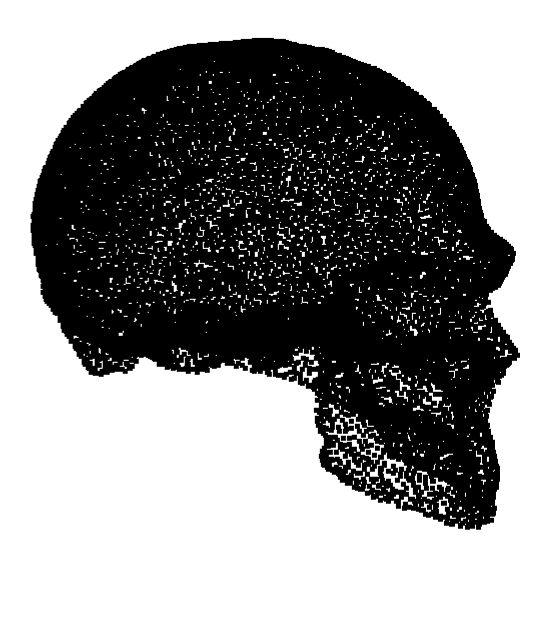}
  \end{minipage} \\ \midrule
Uid & e1f7278ffd7c4878a3cffbc98f05359a & 2b7705b7978f4d7e819a90f914eb3357 \\
Ground Truth & Wooden hammer with wooden handle - 3D model. & A 3D model of a human skull. \\ 
Category & Hammer & Skull \\

\midrule

User & What is this? & This is an object of \\
ULIP-based LLM & A hammer, screwdriver, and wrench. & A sphere with a hole in it. \\
JM3D-LLM & A hammer with a wooden handle. & A low poly model of a human skull. \\ \bottomrule

Samples 9,10 & 
  \begin{minipage}{\linewidth}
    \hspace{1.5em}
    \includegraphics[width=0.36\linewidth]{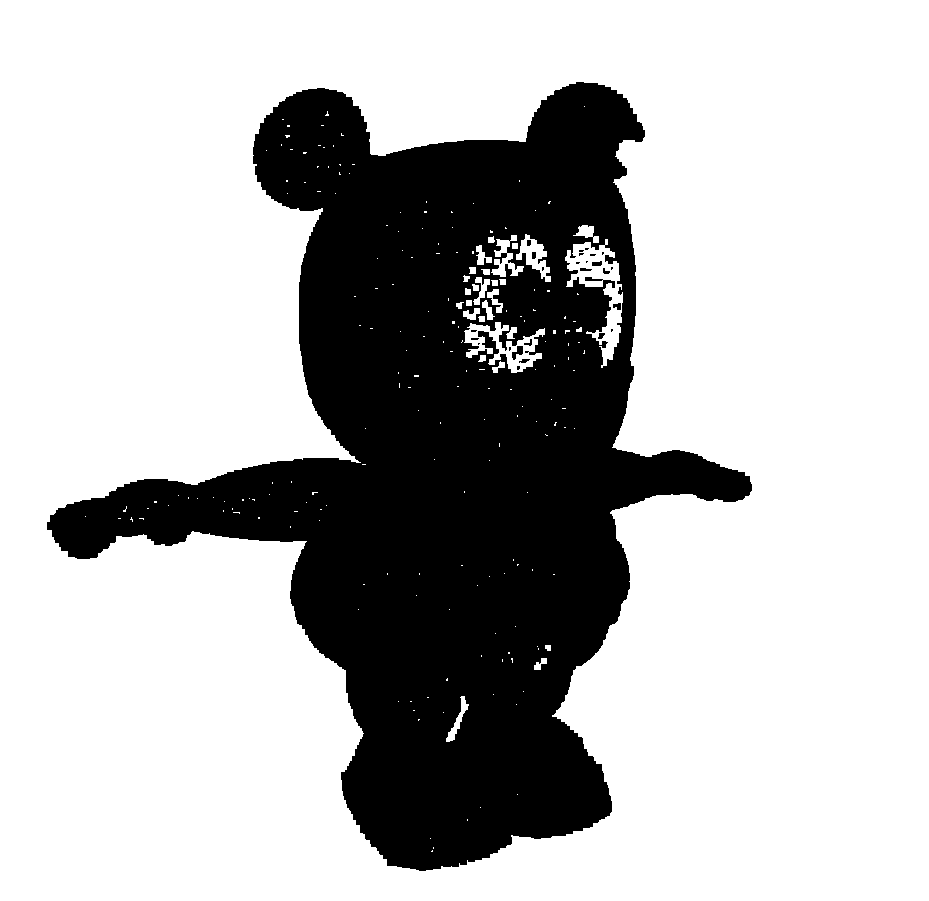}
    \hspace{1.em}
    \includegraphics[width=0.36\linewidth]{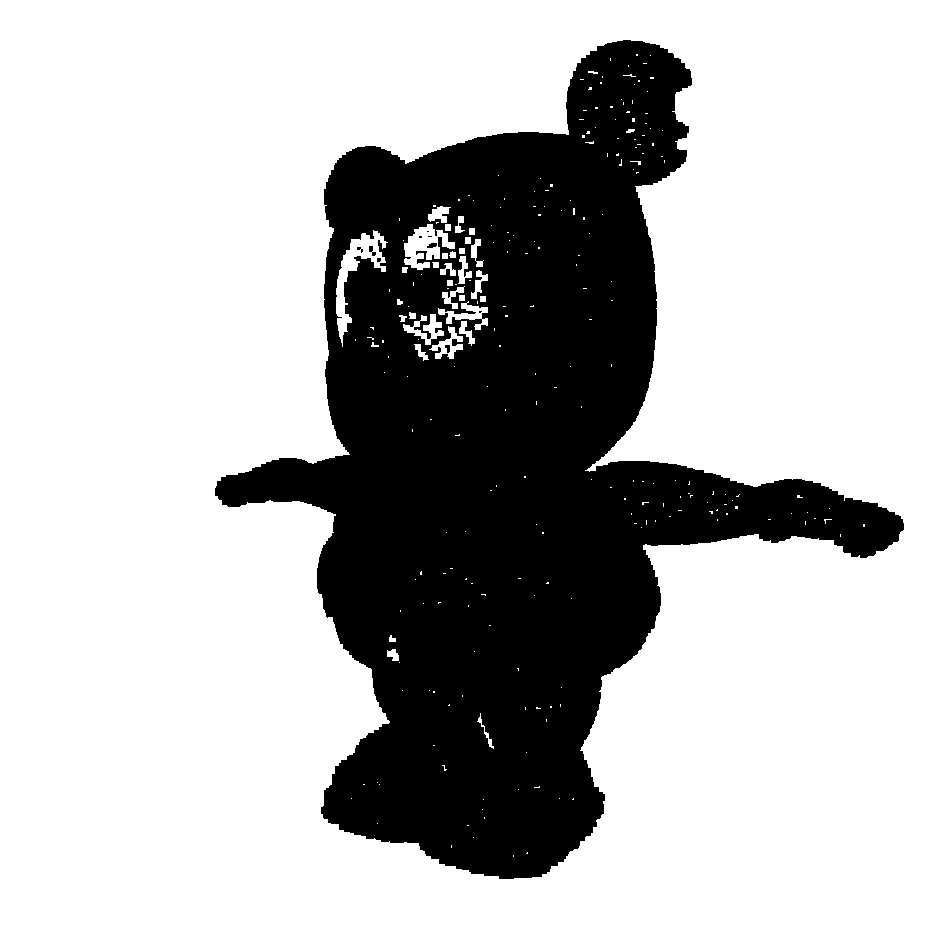}
  \end{minipage}
& 
  \begin{minipage}{\linewidth}
    \hspace{3.35em}
    \includegraphics[width=0.24\linewidth]{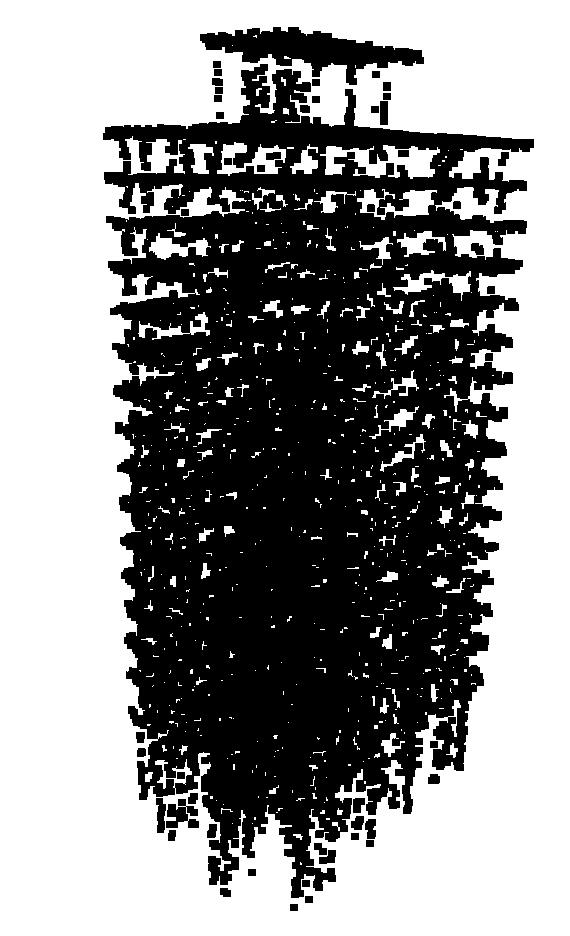}
    \hspace{5.5em}
    \includegraphics[width=0.24\linewidth]{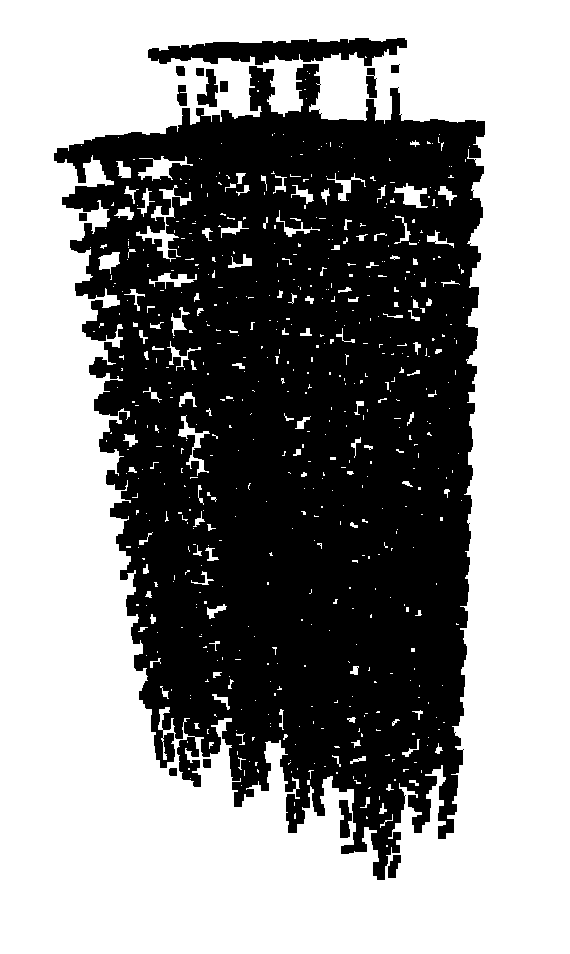}
  \end{minipage} \\ \midrule

Uid & 0fe7d12ef21f422583a873cb4d4fa89f & 69865c89fc7344be8ed5c1a54dbddc20 \\
Ground Truth & Green 3D cartoon bear with arms outstretched and wearing sneakers. & A storyed building. \\ 
Category & Bear & Building \\
\midrule
User & What is it? & What is it? \\
ULIP-based LLM & A cartoon character wearing a helmet. & A building with a roof. \\
JM3D-LLM & A teddy bear. & A large, intricate, multi-layered structure with many lines and wires. \\
\bottomrule

\end{tabular}
}
\label{tab:qualitaive-JM3DLLM}
\end{table*}

\begin{table*}[!h]
\centering
\caption{Qualitative comparisons with ULIP-based LLM and our JM3D-LLM on our ShapeNet benchmark.}

\resizebox{1\textwidth}{!}{
\begin{tabular}{@{}l p{0.4\linewidth} p{0.4\linewidth}}
\toprule

Samples 1,2 & 
  \begin{minipage}{\linewidth}
    \includegraphics[width=0.40\linewidth]{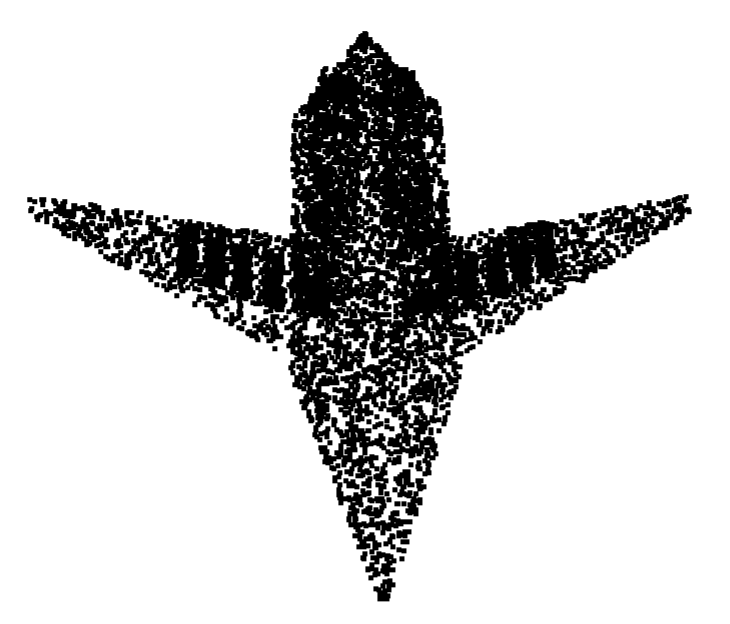}
    \hspace{2.7em}
    \includegraphics[width=0.44\linewidth]{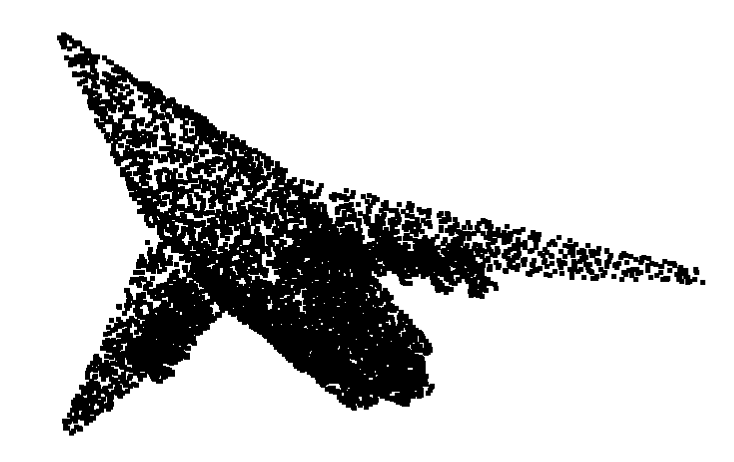}
    \vspace{0.7em}
  \end{minipage}
& 
  \begin{minipage}{\linewidth}
    \hspace{0.8em}
    \includegraphics[width=0.36\linewidth]{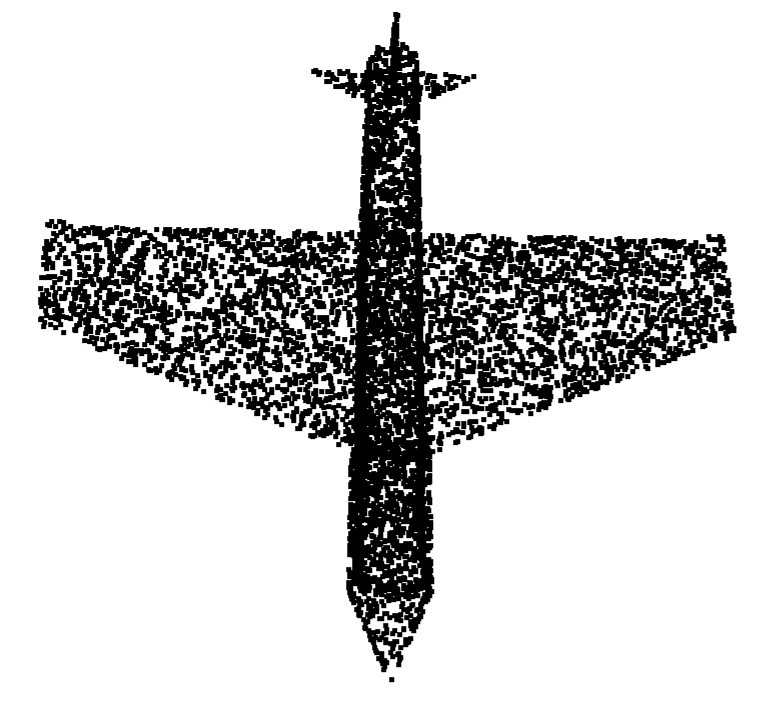}
    \hspace{3.7em}
    \includegraphics[width=0.36\linewidth]{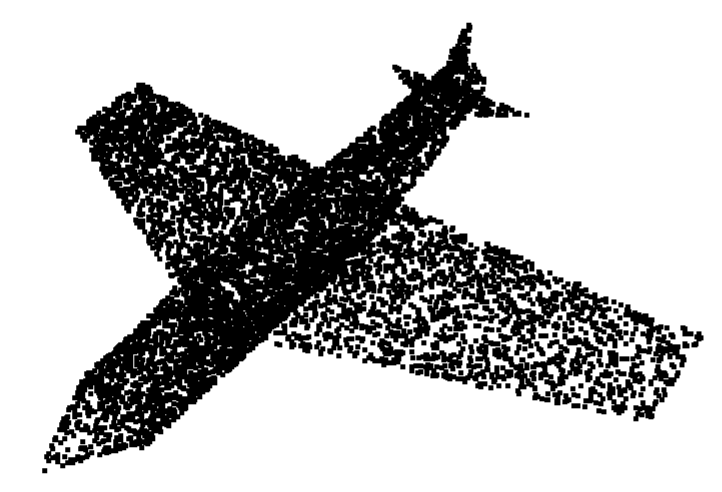}
  \end{minipage}
\\ \midrule
Category & bomber & attack aircraft \\
\midrule

User & What is this? & What is this? \\
ULIP-based LLM & a diorama of a jet-propelled plane. & a diorama of a plane. \\ 
JM3D-LLM & a diorama of a bomber. & a diorama of an attack aircraft. \\ 
\bottomrule

Samples 3,4 & 
  \begin{minipage}{\linewidth}
    \includegraphics[width=0.40\linewidth]{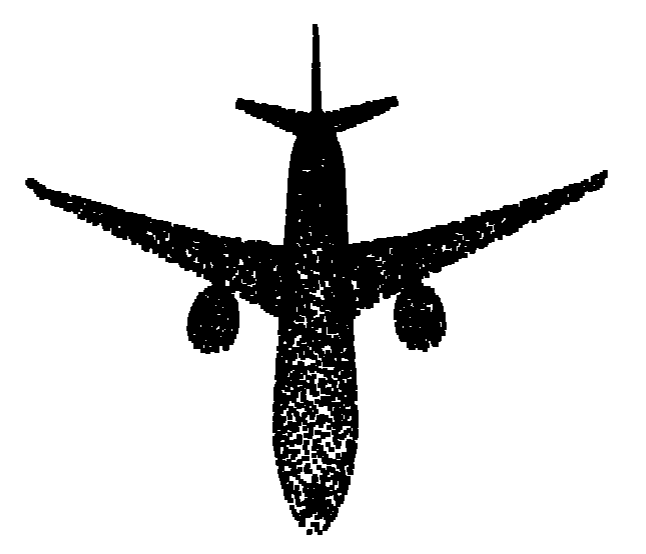}
    \hspace{1.7em}
    \includegraphics[width=0.40\linewidth]{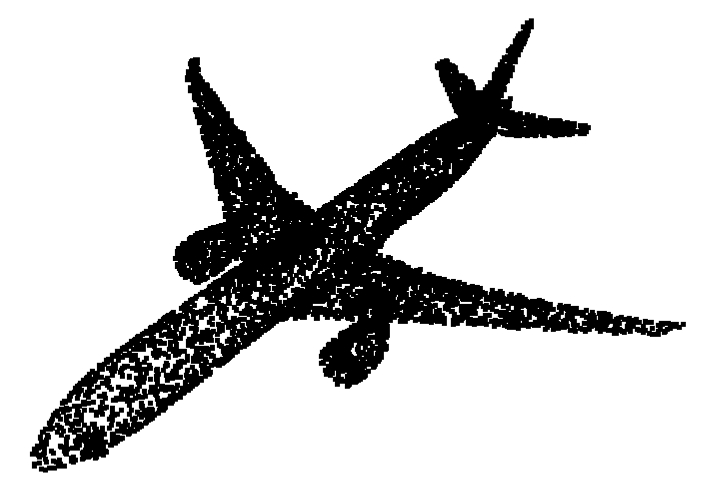}
  \end{minipage}
& 
  \begin{minipage}{\linewidth}
    \includegraphics[width=0.44\linewidth]{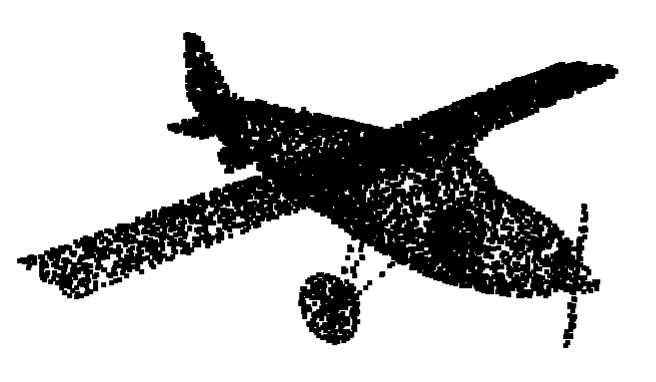}
    \hspace{2.7em}
    \includegraphics[width=0.44\linewidth]{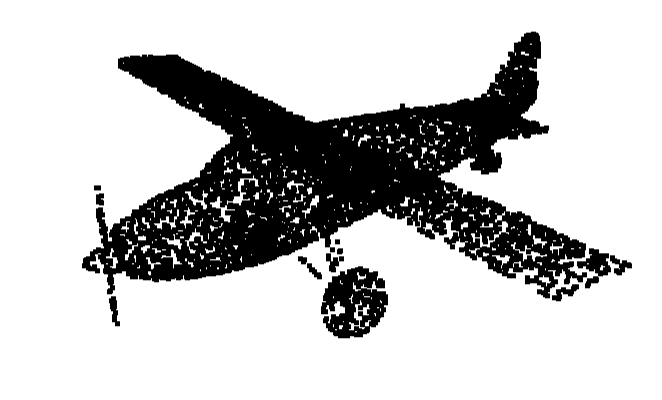}
  \end{minipage} \\ \midrule
Category & jet-propelled plane & propeller plane \\

\midrule

User & What is this? & What is this? \\
ULIP-based LLM & a diorama of a plane. & a diorama of a plane. \\ 
JM3D-LLM & a diorama of a jet-propelled plane. & a diorama of a propeller plane. \\ 
\bottomrule

Samples 5,6 & 
  \begin{minipage}{\linewidth}
    \vspace{1.8em}
    \includegraphics[width=0.40\linewidth]{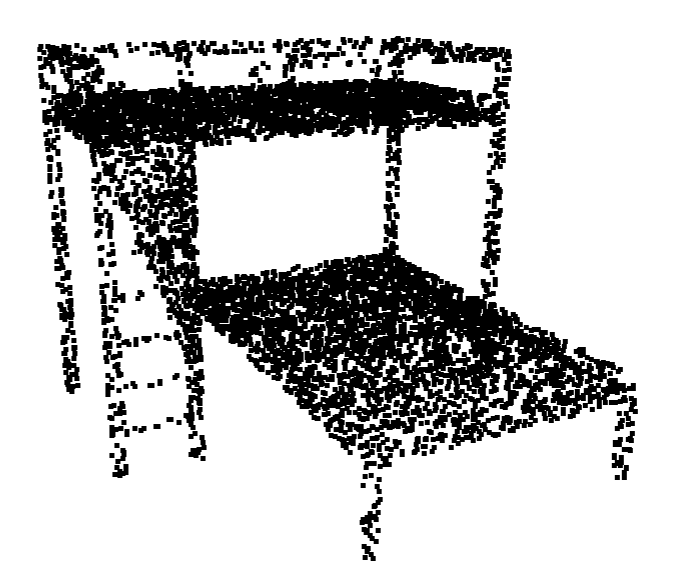}
    \hspace{1.7em}
    \includegraphics[width=0.40\linewidth]{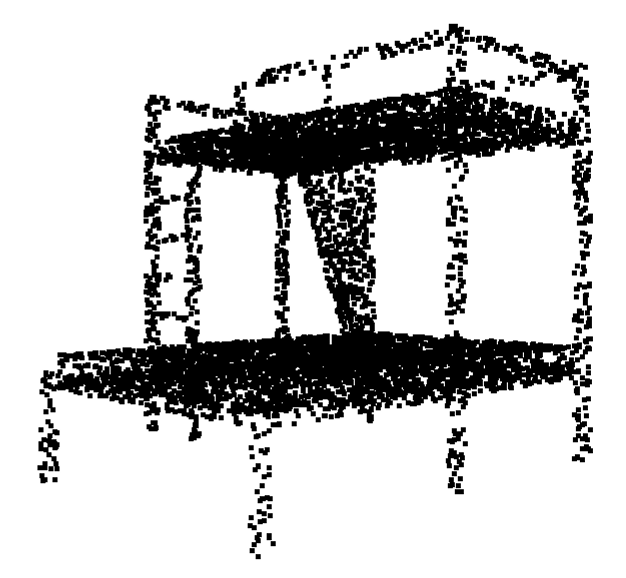}
    \vspace{1.8em}
  \end{minipage}
& 
  \begin{minipage}{\linewidth}
    \includegraphics[width=0.40\linewidth]{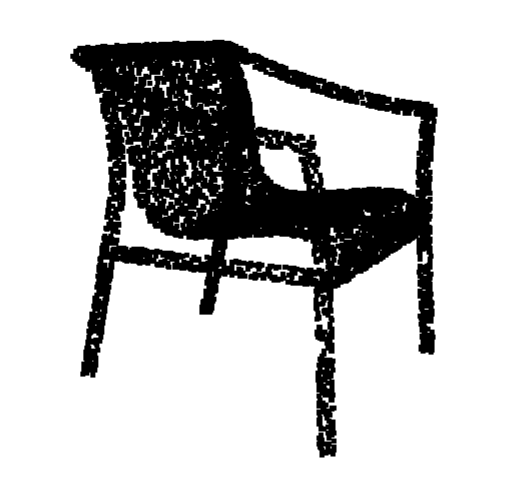}
    \hspace{3.7em}
    \includegraphics[width=0.40\linewidth]{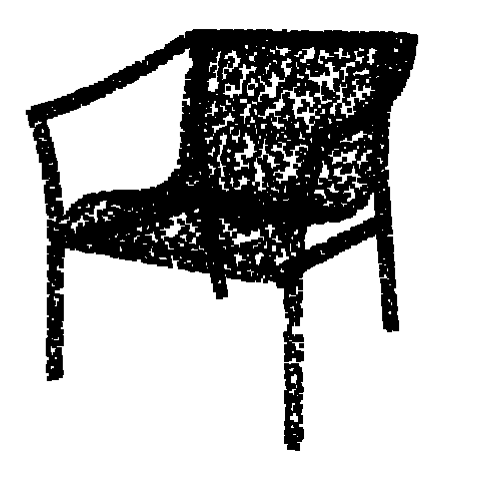}
  \end{minipage} \\ \midrule 
Category & bunk bed & armchair \\
\midrule
User & What is this? & What is this? \\
ULIP-based LLM & a diorama of a bed. & a diorama of a park bench. \\ 
JM3D-LLM & a diorama of a bunk bed. & a diorama of a armchair. \\ 
\bottomrule

Samples 7,8 & 
  \begin{minipage}{\linewidth}
    \vspace{1.8em}
    \includegraphics[width=0.44\linewidth]{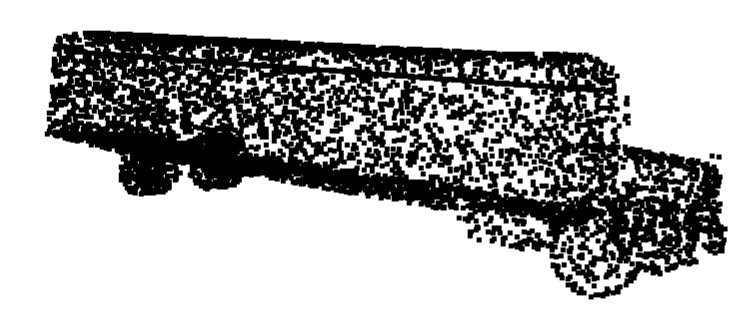}
    \hspace{1.8em}
    \includegraphics[width=0.44\linewidth]{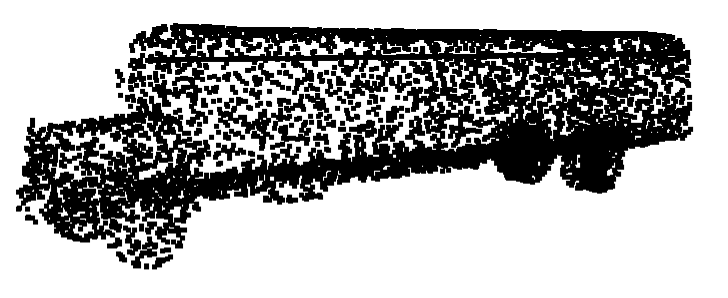}
    \vspace{1.8em}
  \end{minipage}
& 
  \begin{minipage}{\linewidth}
    \hspace{1.8em}
    \includegraphics[width=0.36\linewidth]{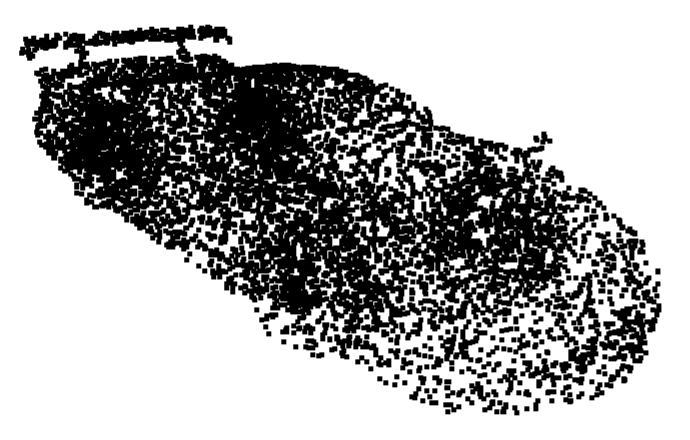}
    \hspace{2.3em}
    \includegraphics[width=0.44\linewidth]{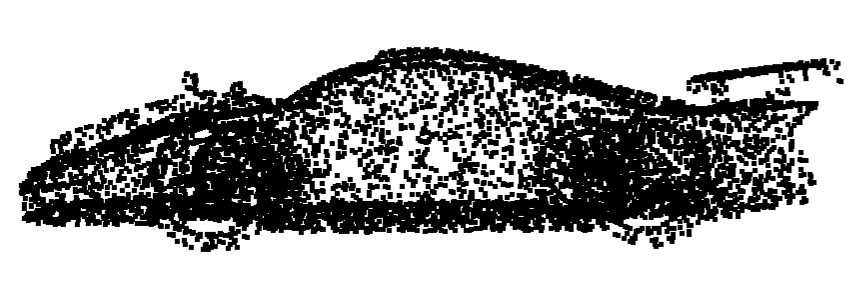}
  \end{minipage} \\ \midrule
Category & school bus & racer \\

\midrule

User & What is this? & What is this? \\
ULIP-based LLM & a diorama of a motorcoach. & a diorma of a sport car. \\
JM3D-LLM & a diorama of a school bus. & a diorama of a racer. \\ \bottomrule

Samples 9,10 & 
  \begin{minipage}{\linewidth}
    \vspace{1.8em}
    \hspace{1.5em}
    \includegraphics[width=0.40\linewidth]{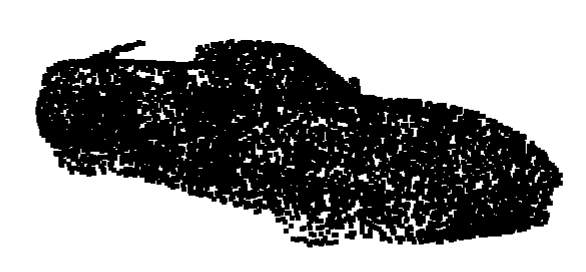}
    \hspace{1.em}
    \includegraphics[width=0.40\linewidth]{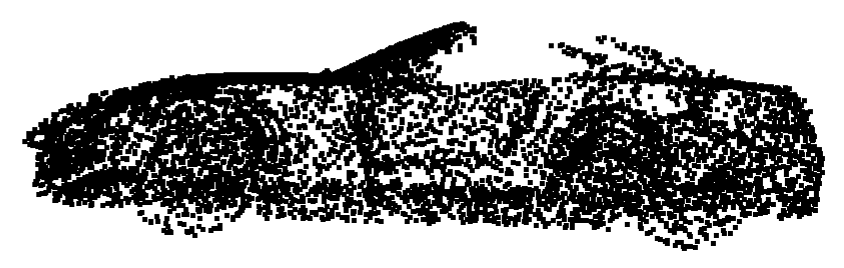}
    \vspace{1.8em}
  \end{minipage}
& 
  \begin{minipage}{\linewidth}
    \hspace{1.35em}
    \includegraphics[width=0.44\linewidth]{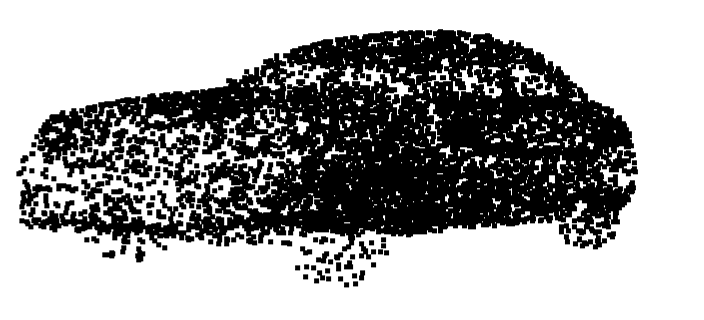}
    \hspace{2.5em}
    \includegraphics[width=0.36\linewidth]{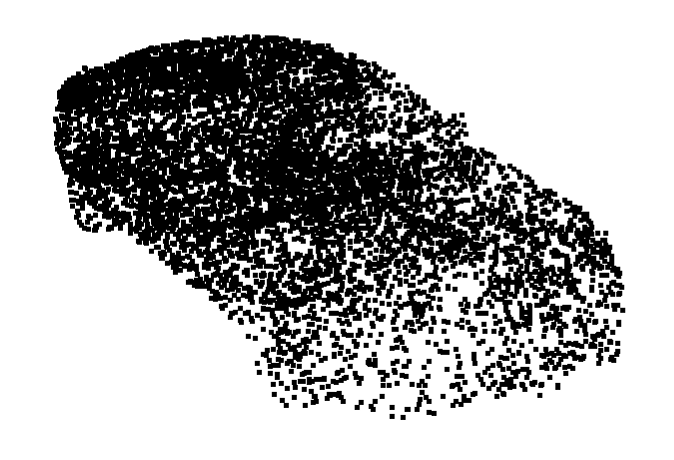}
  \end{minipage} \\ \midrule
  
Category & roadster & saloon \\
\midrule
User & What is it? & What is it? \\
ULIP-based LLM & a diorma of a motorcar in white scene. & a diorama of a coupe. \\
JM3D-LLM & a diorama of a roadster. & a diorama of a saloon. \\
\bottomrule

\end{tabular}
}
\label{tab:qualitaive-JM3DLLM-shapenet}
\end{table*}

\subsection{Semantic and Partial Segmentation Experiments}

We conduct experiments on semantic segmentation using the S3DIS~\cite{armeni20163d} dataset and on partial segmentation with the ShapeNet~\cite{chang2015shapenet} dataset. The results are presented in Tab.~\ref{tab:seg-semantic} for S3DIS and Tab.~\ref{tab:seg-part} for ShapeNet. 
Our proposed method consistently showcases strong performance. Specifically, we achieve scores of 57.1 on the S3DIS dataset and 82.1 on the ShapeNet dataset, marking improvements of 3.6 and 0.3 points respectively. These advancements in scene-understanding tasks unequivocally validate the efficacy of our approach.

\subsection{Exploring Cross-Modal Capabilities of JM3D \label{sec:cross-retrival}}

A salient feature of JM3D is its ability to empower the foundational point cloud model with enhanced cross-modal capabilities. In this section, we delve into JM3D's proficiency in preserving uncommon viewpoint information within models. Specifically, we investigate its performance when using images captured from unconventional viewpoints to retrieve 3D models.

We utilize images from the real-world dataset, Caltech101~\cite{fei2004learning}, aiming to retrieve 3D models from the ModelNet40 test set, a medium-scale dataset comprising over 2.5K models spanning 40 categories. Additionally, we construct several challenging samples characterized by their unique perspectives, which are typically difficult for conventional models to recognize. The top-3 retrieval models based on this experiment are showcased in Fig.~\ref{fig3}.

Observing Fig.~\ref{fig3}, we categorize samples into two classes, namely ``airplane'' and ``laptop''. Each class is further bifurcated into two difficulty levels: simple (top) and challenging (bottom). For simpler samples, most models demonstrate commendable retrieval accuracy. However, for images captured from rarer viewpoints, ULIP struggles to match the correct point cloud. In stark contrast, JM3D trained with two views yields some correct matches. Moreover, when the image count increases to four in the CIS, JM3D almost flawlessly identifies the correct models. These results are a testament to our model's capability to bridge meaningful features between visual and 3D modalities. Furthermore, as indicated by Tab.~\ref{tab:ablation-image}, while increasing the number of views in the CIS might marginally influence text-based performance, it significantly bolsters the model's alignment prowess in the image domain.

\subsection{Detailed Captioning from JM3D-LLM}
While the experiments discussed thus far primarily assess the capabilities of 3D representations in classification tasks, this subsection seeks to explore the representational power of various features in terms of fine-grained cues. To achieve this, we incorporate 3D representations into a Large Language Model (LLM), leveraging the LLM to parse detailed description results. The outcomes are presented in Tab.~\ref{tab:qualitaive-JM3DLLM}. Contrasting with ULIP-based LLM, JM3D-LLM emerges superior in generating more precise descriptions, rather than merely offering a simplistic categorization. As Tab.~\ref{tab:qualitaive-JM3DLLM} reveals, JM3D demonstrates a nuanced understanding not only of category information but also discerns material properties (sample 1), intricate instance details (sample 2 and sample 3), and even abstract concepts like car brands (sample 5). By bridging with the LLM, we can produce descriptions with abstract concepts or more detailed insights—such as the ``teddy bear'' in sample 9 or the comprehensive analysis of house structure in sample 10—that ULIP fails to capture. These findings underscore the formidable linguistic prowess of LLM and, crucially, attest to JM3D's training strategy that meticulously conserves precise fine-grained features.

Furthermore, capitalizing on the diverse granular textual information introduced during training, we utilize JM3D-LLM to delve deeper into the fine-grained representational capabilities of our proposed JM3D. As illustrated in Tab.~\ref{tab:qualitaive-JM3DLLM-shapenet}, JM3D-LLM adeptly details the fine-grained categories of the model, such as ``attack aircraft'' and ``armchair''. This substantiates the efficacy of our introduced Hierarchical Text Tree module.

\subsection{Ablation Study}

To unpack the specific contributions of SMO and JMA during the pre-training phase, we conducted ablation studies on these two modules using PointMLP as our foundational model. Given that the ultimate aim is to achieve alignment between point cloud features and image-text features, we employed zero-shot metrics on ModelNet40 and ScanObjectNN datasets and utilized cross-modal retrieval as our qualitative measure.

\subsubsection{Continuous Multi Views vs. Random One Look}

The results of this comparison are presented in Tab.~\ref{tab:ablation-image}. We initiated the study by examining the effect of varying numbers of viewpoint images. Intriguingly, integrating multiple viewpoints directly results in a performance dip. This suggests that the semantic continuity across different viewpoint features is disrupted abruptly, leading the model to grapple with diverse information stemming from the multiple perspectives. However, it becomes evident that by incorporating embeddings, the model’s proficiency in discerning between viewpoints escalates, accounting for a 2.3\% improvement. Introducing within-view sampling further bolsters the semantic continuity of viewpoints, leading to a 1.9\% boost. On the flip side, an excess of images skews the model towards aligning predominantly with image features. This becomes manifest when performance diminishes as the number of images rises from 2 to 4. Concurrently, the prowess in image retrieval enhances, a phenomenon that is elaborated upon in Sec.\ref{sec:cross-retrival}.

\subsubsection{Hierarchical Text vs. Pre-defined Text}

Following the validation of the efficacy of CIS, we spotlight the significance of the text tree as displayed in Tab.~\ref{tab:ablation-text}. Merely introducing subcategory text results in a negligible uptick – a mere 0.6\% enhancement in top-1 accuracy. This emphasizes that the granularity of the text isn’t the crux of the matter. Yet, the incorporation of the structured category tree, specifically, the HTT, propels a commendable 1.3\% improvement. It's worth mentioning that the top-5 accuracy with subcategories is diminishing. This can be attributed to the augmented categories, which ratchets up the intricacy of aligning linguistic features. Overall, the HTT aids in minimizing the semantic gap between samples clustered under a parent category. By doing so, the HTT infuses structured semantic insights, optimizing the alignment between point cloud and textual features.

\subsubsection{Collaborative Alignment vs. Independent Alignment}

As shown in Tab.~\ref{tab:ablation-joint}, the inclusion of JMA results in a 2.7\% enhancement in top-1 accuracy, underscoring the efficacy of JMA. While SMO provides a more streamlined data organization, it falls short in markedly boosting the model's performance. Aligning 3D representations separately with images and text can lead to unstable optimization. By incorporating JMA, we transition from the rigid assumption of independent alignment in the text-image domain to a more joint modeling approach. This shift significantly strengthens the alignment of point clouds.

\section{Conclusion}

We introduce JM3D, a comprehensive pre-training framework that incorporates both the SMO and JMA modules. This framework harmoniously merges language, image, and point cloud features into a cohesive semantic space, circumventing the need for any specialized architecture. With the precision of the SMO module in harnessing information across various modalities and the JMA module's novel approach to joint modeling, the alignment across these modalities is optimized. Delving deeper, we extend our framework to JM3D-LLM, which synergistically combines 3D representation with large language models through an efficient fine-tuning process. The outstanding results achieved by JM3D in zero-shot 3D classification and image retrieval tasks, establishing a new benchmark, underline its unmatched cross-modal prowess. The precise and rich descriptions provided by JM3D-LLM further attest to the formidable representational capacity of JM3D.





\section*{Acknowledgments}
This work was supported by National Key R\&D Program of China (No.2022ZD0118201), the National Science Fund for Distinguished Young Scholars (No.62025603), the National Natural Science Foundation of China (No. U21B2037, No. U22B2051, No. 62176222, No. 62176223, No. 62176226, No. 62072386, No. 62072387, No. 62072389, No. 62002305 and No. 62272401), the National Natural Science Fund for Young Scholars of China (No. 62302411), China Postdoctoral Science Foundation (No.2023M732948), and the Natural Science Foundation of Fujian Province of China (No.2021J01002,  No.2022J06001).

\ifCLASSOPTIONcaptionsoff
  \newpage
\fi



\bibliographystyle{IEEEtran}
\bibliography{IEEEtranTPAMI}

\end{document}